%% file: ViCGCN.tex
\documentclass[final,1p,times]{elsarticle}

\usepackage{hyperref}
\usepackage[T5]{fontenc} 
\usepackage{latexsym} 
\usepackage{graphicx} 
\usepackage{pgf-pie} 
\usepackage{soul}
\usepackage{color}
\usepackage{tablefootnote}
\usepackage{url}
\usepackage{caption}
\usepackage{float}
\usepackage{subfigure}
\usepackage{tabto}
\usepackage{tabularx}
\usepackage{tabularray}
\usepackage{mathtools}
\usepackage{amsmath}
\usepackage{adjustbox}
\usepackage{lscape}
\usepackage{amsfonts}
\usepackage{multirow}

\usepackage{comment}

\begin{document}
\let\WriteBookmarks\relax
\def\floatpagepagefraction{1}
\def\textpagefraction{.001}

\begin{frontmatter}

\title{ViCGCN: Graph Convolutional Network with Contextualized Language Models for Social Media Mining in Vietnamese}


\author[label1,label2]{Chau-Thang~Phan}
\ead{20520929@gm.uit.edu.vn}    
\author[label1,label2]{Quoc-Nam~Nguyen}
\ead{20520644@gm.uit.edu.vn}
\author[label1,label2]{Chi-Thanh~Dang}
\ead{20520761@gm.uit.edu.vn}
\author[label1,label2]{Trong-Hop~Do}
\ead{hopdt@uit.edu.vn}
\author[label1,label2]{Kiet~Van~Nguyen\corref{cor1}}
\ead{kietnv@uit.edu.vn}
\cortext[cor1]{Corresponding author at the University of Information Technology, Vietnam National University, Ho Chi Minh City, Vietnam.}

\affiliation[label1]{organization={Faculty of Information Science and Engineering, University of Information Technology},
            city={Ho Chi Minh city},
            country={Vietnam}}
\affiliation[label2]{organization={Vietnam National University},
            city={Ho Chi Minh city},
            country={Vietnam}}

\journal{Neurocomputing}

\begin{abstract}
Social media processing is a fundamental task in natural language processing (NLP) with numerous applications. As Vietnamese social media and information science have grown rapidly, the necessity of information-based mining on Vietnamese social media has become crucial. However, state-of-the-art research faces several significant drawbacks, including imbalanced data and noisy data on social media platforms. Imbalanced and noisy are two essential issues that need to be addressed in Vietnamese social media texts. Graph Convolutional Networks can address the problems of imbalanced and noisy data in text classification on social media by taking advantage of the graph structure of the data. This study presents a novel approach based on contextualized language model (PhoBERT) and graph-based method (Graph Convolutional Networks). In particular, the proposed approach, ViCGCN, jointly trained the power of \textbf{C}ontextualized embeddings with the ability of Graph Convolutional Networks, \textbf{GCN}, to capture more syntactic and semantic dependencies to address those drawbacks. Extensive experiments on various Vietnamese benchmark datasets were conducted to verify our approach. The observation shows that applying GCN to BERTology models as the final layer significantly improves performance. Moreover, the experiments demonstrate that ViCGCN outperforms 13 powerful baseline models, including BERTology models, fusion BERTology and GCN models, other baselines, and state-of-the-art methods on three benchmark social media datasets. Our proposed ViCGCN approach demonstrates a significant improvement of up to 6.21\%, 4.61\%, and 2.63\% over the best Contextualized Language Models, including multilingual and monolingual, on three benchmark datasets, UIT-VSMEC (for Vietnamese emotion recognition), UIT-ViCTSD (for Vietnamese constructive and toxic analysis), and UIT-VSFC (for Vietnamese sentiment analysis), respectively. Additionally, our integrated model ViCGCN achieves the best performance compared to other BERTology integrated with GCN models. The code\footnote{\url{https://github.com/phanchauthang/ViCGCN}} is publicly available for research purposes.
\end{abstract}






\end{frontmatter}



\input{sections/intro}
\input{sections/related}
\input{sections/vicgcn}
\input{sections/experiments}
\input{sections/conclusion}





\bibliographystyle{elsarticle-num-names} 
\bibliography{cas-refs.bib}


\end{document}

%% file: sections/intro.tex
\section{Introduction}
\label{introduction}

Social media processing is the essential task in natural language processing (NLP) \cite{Natural-Language-Processing-for-Social-Media, injadat2016data}, utilized in many world applications over the last few decades, such as topic detection \cite{topic-detection}, spam detection \cite{Spam-detection-in-twitter}, opinion mining \cite{Opinion-Mining-on-Social-Media-Data}, and text classification \cite{A-survey-on-text-mining-in-social-networks} in general. Along with information science and engineering development, the world has witnessed exponential growth in social media platforms in many countries, including Vietnam. According to the Digital 2021 report\footnote{\url{https://datareportal.com/reports/digital-2021-global-overview-report}} by We Are Social and Hootsuite (2021), there were 4.2 billion active social media users worldwide in January 2021, an increase of 13\% compared to the previous year. Facebook\footnote{\url{https://www.facebook.com/}} alone had 2.7 billion active users in the same month, making it the largest social media platform in the world. Instagram\footnote{\url{https://www.instagram.com/}}, owned by Facebook, had over 1 billion active users in 2020, while Twitter\footnote{\url{https://twitter.com/}} had 353 million monthly active users in the same year. Furthermore, the report shows that people spend an average of 2 hours and 25 minutes daily on social media and messaging apps. These statistics show social media platforms' widespread use and growing popularity worldwide. Therefore, social media platforms (e.g., Facebook\footnote{\url{https://www.facebook.com/}}, Twitter\footnote{\url{https://twitter.com/}}, and Instagram\footnote{\url{https://www.instagram.com/}}) have become an integral part of modern communication and is a rich source of information for various applications, including sentiment analysis, opinion mining, and recommendation systems. However, social media mining in Vietnamese poses unique challenges due to the complexity of the language and the informal nature of social media text. Furthermore, Vietnamese social media text often includes slang, misspellings, and other non-standard language features, making applying traditional natural language processing (NLP) techniques difficult. Therefore, the need for effective Social media processing techniques on Vietnamese social media has become more crucial to improving their performance.

According to the essentials of Social media processing on social media, several pieces of research have been studied. In Vietnamese particular, \citet{VSFC} presented a Vietnamese Students’ Feedback Corpus (UIT-VSFC), a free and high-quality corpus with two distinct tasks, namely sentiment-based and topic-based classifications. \citet{DBLP:journals/corr/abs-1911-09339} introduced a standard Vietnamese Social Media Emotion Corpus (UIT-VSMEC), contributing to emotion recognition in Vietnamese. \citet{DBLP:journals/corr/abs-2103-10069} proposed a novel dataset for identifying the constructiveness and toxicity of Vietnamese comments on social media, the Vietnamese Constructive and Toxic Speech Detection dataset (UIT-ViCTSD). A range of Social media processing worked on these Vietnamese social media datasets \cite{huynh-etal-2020-simple,https://doi.org/10.48550/arxiv.2209.10482,Doan_2022,nguyen2020exploiting}. However, these tasks need to be further improved due to the limitations of the models. Furthermore, because of the complexity of social media comments, the current field still faces many challenges due to social media comments' colossal volume and variety in both immensity and topics. One of the primary challenges is dealing with imbalanced data. Social media platforms generate massive amounts of data, often not equally distributed across all classes. This results in an imbalanced dataset, where several classes may have significantly fewer examples than others. Another challenge is handling noisy data, which is common in social media. Social media data often contain many emojis, abbreviations, and non-standard language, making it difficult for Social media processing models to classify the text accurately. Social media comments may also contain sarcasm and irony, which can be challenging to interpret.

Graph Convolutional Networks (GCN) can address the issues of imbalanced and noisy data in Social media processing by taking advantage of the graph structure of the data. By incorporating the graph structure into the model, GCN can effectively capture the relationships and dependencies among the data, which helps to reduce the impact of noise and imbalance. Additionally, Contextualized Language Models can be fine-tuned on specific tasks like text classification, which enables them to adapt to the specific characteristics of the data, such as imbalanced or noisy data. Therefore, our research contributes to text classification on social media in Vietnamese. Our research proposes a novel approach integrating Graph Convolutional Networks with the Vietnamese state-of-the-art Contextualized Language Model, PhoBERT. Our research can handle these issues, including noisy and imbalanced datasets. 

The primary contributions of this study can be outlined as follows: We proposed a novel Vietnamese text classification model ViCGCN by jointly training the large-scale language pre-trained language model PhoBERT and Graph Convolutional Networks (GCN) modules for evaluating social media text classification or social media mining in Vietnamese. Various experiments were conducted with two approaches: BERTology and its Integrated model (BERTology Integrated with GCN), including the Vietnamese state-of-the-art model PhoBERT on three Vietnamese benchmark datasets. Compared to the Vietnamese state-of-the-art PhoBERT model and previous studies done on the datasets, our integrated model ViCGCN achieves significantly better performances. A survey on single large-scale and integrated models was carried out to demonstrate the efficacy of GCN on large-scale pre-trained language models. Additionally, our integrated model ViCGCN successfully addressed the imbalanced and noisy problem of social media datasets. Finally, jointly training BERTology and GCN modules significantly improves performances on Vietnamese social-domain text classification tasks.

The rest of this paper is structured in the following manner. Section \ref{Related work} surveys several current works on Vietnamese text classification. Our proposed approach is presented in detail through Section \ref{Proposed model}. Section \ref{Experiments} briefly looks at the used datasets for our experiments, illustrates processes for implementing models and our experimental results on each task, and describes the result analysis and discussion of the proposed approach. In summary, Section \ref{Conclusion} serves as the final part of our research and outlines our conclusions and any potential areas for future exploration.

%% file: sections/related.tex
\section{Background and Related Work} \label{Related work}
To provide a comprehensive survey related to our proposed model, we review three emerging techniques:  BERTology models, Graph Convolutional Networks, and fusion BERTology models and Graph Convolutional Networks, respectively.

\textbf{BERTology (BERT and its variants):} In natural language processing (NLP), the main models for sequence transduction rely on complex recurrent neural networks \cite{LSTM} or convolutional neural networks \cite{Empirical-Evaluation-of-Gated-Recurrent-Neural-Networks-on-Sequence-Modeling}. However, the sequential nature of these models makes it infeasible to parallelize them when processing longer sequences, which poses a significant challenge. This problem is due to memory constraints that limit batching across examples. As a solution, \citet{AttentionIsAllYouNeed} proposed a new architecture called Transformer that uses attention mechanisms exclusively, eliminating the need for recurrence and convolutions. Inspired and inherited by Transformer, \citet{devlin-etal-2019-bert} proposed the idea of Bidirectional Encoder Representations from Transformers (BERT). This pre-trained model utilizes masked language models to create deep bidirectional representations. BERT obtains new state-of-the-art results on eleven NLP tasks upon launch, including nine GLUE tasks\footnote{\url{https://huggingface.co/datasets/glue}}, SQuAD\footnote{\url{https://huggingface.co/datasets/squad}} v1.0 and 2.0, and SWAG\footnote{\url{https://huggingface.co/datasets/swag}}. Following the success of BERT, several variants of it have been proposed and achieved improved results compared to BERT. Shortly after the release of BERT, mBERT was published by \citet{devlin-etal-2019-bert}. mBERT provides sentence representations for 104 languages. BERT and its variant were called as BERTology \cite{rogers2021primer}. RoBERTa \cite{DBLP:journals/corr/abs-1907-11692}, an optimized version of BERT, which is trained on a larger dataset, was proposed. Aside from multilingual versions or BERT-based models, researchers from different countries are promoted to build and improve monolingual models based on available BERT architectures for their languages: MacBERT for Chinese \cite{Chinese-BERT}, CamemBERT for France \cite{CamemBERT}. PhoBERT \cite{nguyen-tuan-nguyen-2020-phobert} was the initial public large-scale monolingual language model pre-trained for Vietnamese. PhoBERT provides a consistent improvement over the best pre-trained multilingual model, demonstrating superior performance in several Vietnamese-specific NLP tasks and advancing the state-of-the-art in this field.    

\textbf{Graph Neural Networks:} Recently, interest in Graph Neural Networks (GNNs) has grown \cite{wu2020comprehensive}. Representative examples of GNNs proposed by the present include Graph Convolutional Networks (GCN) and its variants, which is one of the most prominent graph deep learning models \cite{Graph-convolutional-networks}. \citet{semi-supervised} presented an algorithm with a GCN for semi-supervised node classification and achieved state-of-the-art classification results on several network datasets. Since then, GCN has been utilized in various applications, to name a few, prediction tasks \cite{prediction3, prediction4, prediction5}, recommendation tasks \cite{recommender1, recommender2}, and classification tasks \cite{class3, class4, class5}. Especially in the field of NLP, GCN has successfully explored NLP tasks such as semantic role labeling (\cite{Encoding-Sentences-with-Graph-Convolutional-Networks-for-Semantic-Role-Labeling, Adaptive-Convolution-for-Semantic-Role-Labeling}, machine translation \cite{Graph-Convolutional-Encoders-for-Syntax-aware-Neural-Machine-Translation, Exploiting-Semantics-in-Neural-Machine-Translation-with-Graph-Convolutional-Networks}, information extraction \cite{GraphIE, Cardinal-Graph-Convolution-Framework-for-Document-Information-Extraction}, relation extraction \cite{Graph-Convolution-over-Pruned-Dependency-Trees-Improves-Relation-Extraction, Attention-Guided-Graph-Convolutional-Networks-for-Relation-Extraction, Dual-Graph-Convolutional-Networks-for-Graph-Based-Semi-Supervised-Classification}, and text classification also. In the context of text classification, several GCN models have proposed, to name a few: An extension of the GCN framework to the inductive setting called GraphSAGE that allows embeddings to be efficiently generated for unseen nodes in \cite{Inductive-Representation-Learning-on-Large-Graphs}. \citet{FastGCN} presented FastGCN, a fast improvement of the GCN model for learning graph embeddings. FastGCN achieves significant speedups compared to traditional GCN training methods while maintaining similar or even better performance on several benchmark datasets. GCN recursively aggregates neighbor node representations, causing each layer's receptive field to grow exponentially. To address this drawback, \citet{Stochastic-Training-of-Graph-Convolutional-Networks-with-Variance-Reduction} introduced algorithms based on control variates to decrease the size of the receptive field. In \cite{Dual-Graph-Convolutional-Networks-for-Graph-Based-Semi-Supervised-Classification}, the authors proposed Dual Graph Convolutional Networks (DGCN) as a simple and scalable method for semi-supervised learning on graph-structured data which can be applied when only a tiny portion of the training data is labeled. \citet{Learning-Graph-Pooling-and-Hybrid-Convolutional-Operations-for-Text-Representations} proposed a graph pooling layer and the hybrid convolutional (hConv) layer that integrates GCN and regular convolutional operations to incorporate node ordering information, achieving a better performance over the conventional CNN-based and GCN-based methods. Recognizing the inherent complexity and redundant computations in GCN, \citet{Simplifying-Graph-Convolutional-Networks} proposed Simplifying Graph Convolutional Networks (SGCN) as a solution to address these issues. In a separate research endeavor, \citet{Graph-Convolutional-Networks-for-Text-Classification} focused on constructing a heterogeneous graph using a corpus and introduced a model known as Text Graph Convolutional Networks (TextGCN). This model employs graph neural networks to concurrently learn word and document embeddings, and it significantly outperformed existing state-of-the-art methods across multiple benchmark datasets.


\textbf{Graph Neural Networks Integrated with BERT:} Following the success of BERT, Graph Neural Networks, and their variants, researchers proposed new models by applying them together. \citet{Graph-Bert} proposed GRAPH-BERT (Graph-Based BERT), which relies entirely on the attention mechanism for graph representation learning without any graph convolution or aggregation operators. \citet{VGCN-BERT} presented the VGCN-BERT model, which integrates the power of BERT with a Vocabulary Graph Convolutional Network (VGCN) to capture global information about the language vocabulary. \citet{Bert-Enhanced} proposed the Bert-Enhanced text Graph Neural Network model (BEGNN), in which text features are extracted using GNN, while semantic features are extracted using BERT. \citet{BertGCN} introduced Bert-GCN, which merges transductive learning with extensive pre-training to accomplish text classification.

%% file: sections/vicgcn.tex
\section{ViCGCN:  Contextualized Graph Convolution Networks for Vietnamese Social Media}
\label{Proposed model}

\begin{figure}[!ht] 
     \centering
     \includegraphics[width=\textwidth]{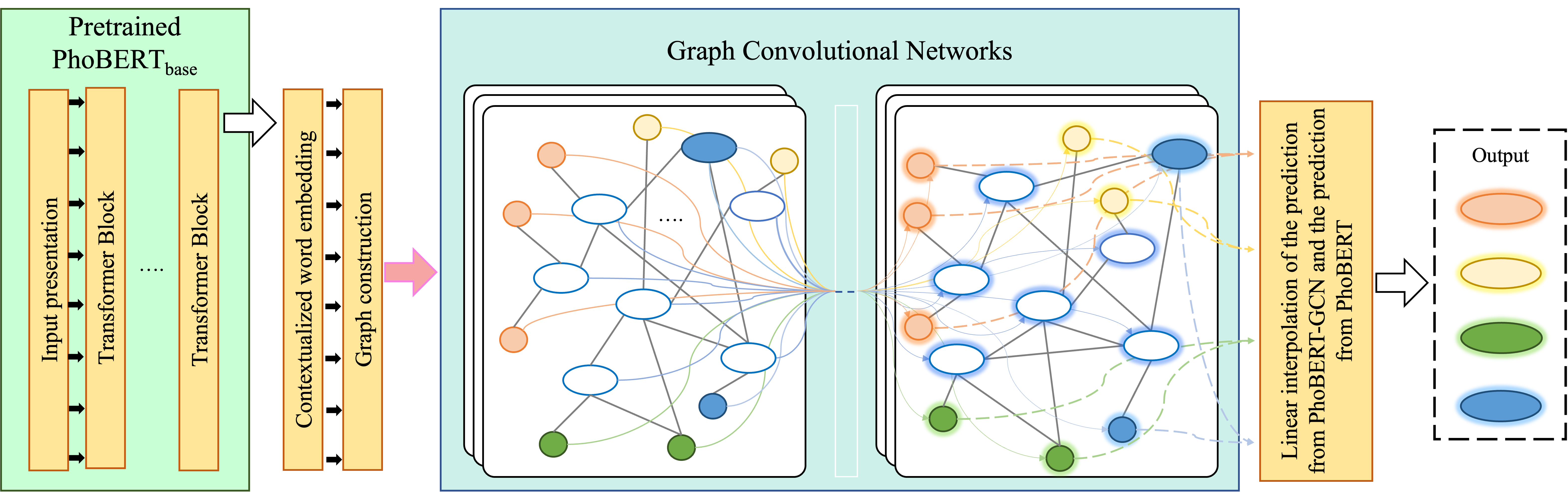}
     \caption{An overview of our proposed approach ViCGCN architecture.}
     \label{fig::ProposedModel/PhoBERT-GCN}
 \end{figure}

Contextualized language models like BERT have shown impressive performance in a wide range of NLP tasks, especially in tasks that require a deep understanding of the meaning of language, such as text classification, sentiment analysis, and named entity recognition. The explanation for this phenomenon is contextual models can capture the contextual meaning of words based on their surrounding words, which is crucial for many NLP tasks. On the other hand, Graph Convolutional Networks (GCN) is a type of graph neural network that can handle graph-structured data, such as text-based dependency graphs, commonly used in Vietnamese language processing. Additionally, GCN is more suitable for semi-supervised learning tasks where the training data is limited and noisy. As a result, the combination of contextualized language model and GCN allows for better modeling of the text data, capturing the complex relationships between words and sentences in a text corpus, leading to improved or showed state-of-the-art (SOTA) performance on a variety of NLP tasks. In this study, we proposed ViCGCN Integrated model and evaluated its efficacy in social media processing for Vietnamese. The ViCGCN architecture consists of two layers, namely the PhoBERT layer and the GCN layer, respectively. Figure \ref{fig::ProposedModel/PhoBERT-GCN} presents an overview of our proposed approach's architecture.

Firstly, we present the architecture of PhoBERT\footnote{\url{https://huggingface.co/vinai/phobert-base}} (\cite{nguyen-tuan-nguyen-2020-phobert}) and how the PhoBERT model performs as the first layer of our proposed approach. PhoBERT was chosen because PhoBERT is specifically designed for the Vietnamese language, making it highly effective for Vietnamese language processing tasks. 
PhoBERT architecture is based on the popular Bidirectional Encoder Representations from Transformers\footnote{\url{https://github.com/google-research/bert}}, also called BERT, architecture, which uses a transformer network to encode the input text and generate high-quality representations of the text.

\begin{figure}[!ht]
    \centering
    \includegraphics[width=\textwidth]{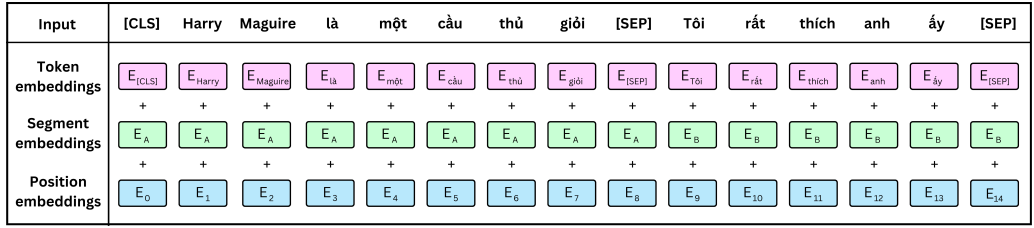}
    \caption{The process of representing the contextualized language model's input. "Harry Maguire là một cầu thủ giỏi. Tôi rất thích anh ấy" is "Harry Maguire is a good player. I very like him" in English.}
    \label{fig::Proposed/PhoBERT}
\end{figure}

The input to these layers is tokenized text, which is then converted into embeddings using the embedding layer as illustrated in Figure \ref{fig::Proposed/PhoBERT}. These embeddings are then processed through the transformer blocks to generate contextualized word representation. In addition to the transformer layers, PhoBERT also includes a pre-processing layer which is responsible for tokenization, sentence segmentation, and special token handling. In this study, PhoBERT is accountable for processing the input text. It takes in the raw text input and applies a series of transformer-based layers. This produces a contextualized embedding for each word in the input. Then, these contextualized embeddings are fed into the GCN layer. The output of the PhoBERT layer represents the contextualized embeddings for each word in the input.

\begin{figure}[!htb]
    \centering
    \includegraphics[width=\textwidth]{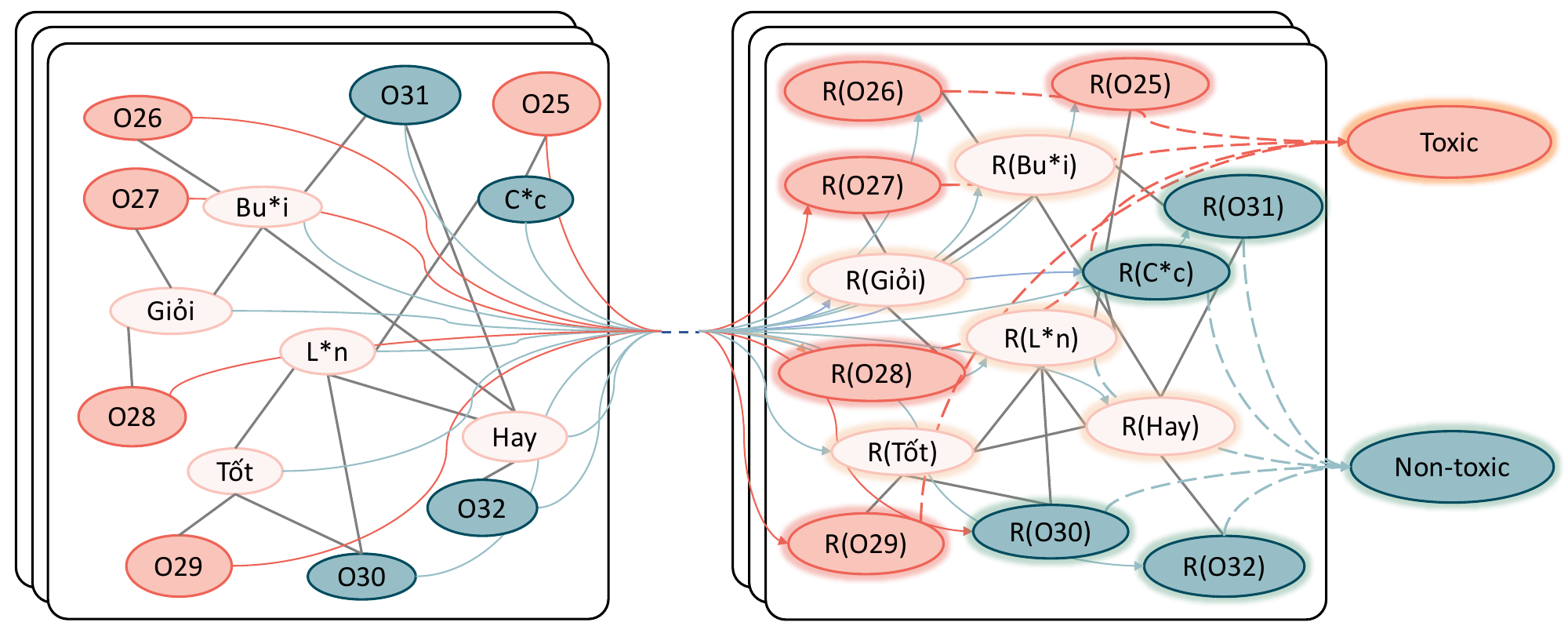}
    \caption{Schematic of GCN layer in ViCGCN. A social media example is taken from UIT-ViCTSD dataset. "Bu*i", "C*c", "Giỏi", "Hay", and "Tốt" are "D*ck", "C*ck", "excellent", "Good", and "Nice" in English, respectively.}
    \label{fig::Proposed/TextGCN}
\end{figure}

 The second layer, the GCN layer, on the other hand, takes the output of the BERT layer, which is a sequence of contextualized word embeddings, as input, and applies graph convolution operations to aggregate information from the surrounding words in a sentence. To be more specific, we create a heterogeneous graph that comprises both document nodes and word nodes, following the TextGCN\footnote{\url{https://github.com/yao8839836/text_gcn}} \cite{Graph-Convolutional-Networks-for-Text-Classification}. Figure \ref{fig::Proposed/TextGCN} schematically presents the overall GCN layer of our integrated model.

To provide a more comprehensive understanding, the GCN layer in our approach utilizes the dependency parse tree of the sentence to create a graphical representation. In this graph, the sentences' words are represented as nodes, and their syntactic relationships are captured as edges. Within the ViCGCN model, this graph serves to illustrate the relationships among words and sentences within a given text document.

To delve deeper into the methodology, we establish a diverse graph that encompasses nodes representing both words and entire documents, drawing inspiration from TextGCN\footnote{\url{https://github.com/yao8839836/text_gcn}} \cite{Graph-Convolutional-Networks-for-Text-Classification}. To establish connections between word and document nodes, we make use of the term frequency-inverse document frequency (TF-IDF) measure, which helps define the associations between word-document pairs. Additionally, we employ positive point-wise mutual information (PPMI) to establish connections between word-word pairs. The weight of an edge connecting two nodes, denoted as \textit{i} and \textit{j}, is defined as follows:
\begin{equation}
    A_{i, j} = 
    \begin{cases}
        PPMI(i, j), & \textit{i, j are words and i} \neq j \\
        TF-IDF(i, j), & \textit{i is document, j is word} \\
        1, & i = j \\
        0, & otherwise
    \end{cases}
\end{equation}


In ViCGCN, the contextual language model PhoBERT is responsible for acquiring document embeddings and considering them input representations for document nodes. These document node embeddings are represented as $X_{doc} \in \mathbb{R}^{n_{doc} \times d}$, where $n_{doc}$ signifies the count of document nodes, $n_{word}$ represents the count of word nodes (comprising both training and testing), and $d$ denotes the dimensionality of the embeddings. Consequently, the initial matrix of node features is formulated as follows:

\begin{equation}
    X = 
    \begin{pmatrix}
        X_{doc} \\
        0
    \end{pmatrix}_{(n_{doc}+n_{word}) \times d}
\end{equation}


 Then X is fed into a series of Graph Convolutional Networks layers, where each layer aggregates information from the neighbors of each node to refine its representation. More precisely, the output feature matrix for the $i$-th GCN layer, denoted as $L^{(i)}$, is calculated as follows: it involves an activation function represented by $f$, utilizes the normalized adjacency matrix denoted as $\tilde{A}$, and incorporates a weight matrix $W^{(i)} \in \mathbb{R}^{d_{i-1} \times d_i}$ specific to that layer. The initial input feature matrix of the model is denoted as $L^{(0)} = X$.

\begin{equation}
    L^{(i)} = f(\tilde{A}L^{(i-1)}W^{(i})
\end{equation}
 
 The output of the GCN layer is a set of updated embeddings, which capture the interactions among the words in the sentence and pass through a softmax activation layer to obtain the final predictions, where $g$ represents the GCN model: 
 \begin{equation}
     \textbf{Z}_{\text{GCN}} = softmax(g(X, A))
 \end{equation}

Moreover, an auxiliary classifier on BERT embeddings is conducted by directly feeding document embeddings (denoted by $X$) to a dense layer with softmax activation.
\begin{equation}
    \textbf{Z}_{\text{BERT}} = softmax(WX)
\end{equation}

 To combine the output embeddings of the PhoBERT and GCN layers and obtain the best classification performance, we propose to use a hyperparameter $\lambda$ to control the trade-off between them in the final classification. Specifically, we compute a weighted sum of the two embeddings using the following equation:
\begin{equation}
    \mathbf{Z} = \lambda \mathbf{Z}_{\text{GCN}} + (1-\lambda) \mathbf{Z}_{\text{PhoBERT}}
    \label{equa::lambda}
\end{equation}

 where $\mathbf{Z}_{\text{GCN}}$ is the output embedding of the GCN layer and $\mathbf{Z}_{\text{PhoBERT}}$ is the output embedding of the PhoBERT layer. The softmax function normalizes the output and produces class probabilities for text classification. Moreover, comprehensive experiments were conducted on the three benchmarks UIT-VSMEC, UIT-ViCTSD, and UIT-VSFC to determine the optimal lambda value for the ViCGCN model in Section \ref{imapactGCN}.

 By combining the power of PhoBERT's contextualized embeddings with the ability of GCN to capture syntactic and semantic dependencies, the ViCGCN can achieve better performance on social media processing tasks, especially those that require an understanding of semantic relationships between words. Furthermore, the ViCGCN model can also handle a broader range of text inputs, including more extended and more complex sentences, due to its ability to capture the contextualized meaning of words and the syntactic and semantic dependencies between them. This makes it a highly effective tool for natural language processing tasks, especially text classification and social media processing tasks.

%% file: sections/experiments.tex
\section{Experiments and Analysis} \label{Experiments}
\subsection{Experimental Design}

This section delineates our approach for introducing a novel classification model called ViCGCN. Initially, we gather three benchmark datasets, as mentioned in Section \ref{Experiments/Datasets}, and subject them to a cleaning process described subsequently. Subsequently, the pre-processed data is employed for training both our baseline models and the proposed model. We fine-tune each model to identify optimal hyperparameters and enhance their performance. The evaluation of model performance is conducted using Macro F1-score and Weighted F1-score, as discussed in Section \ref{Experiments/Metrics}. Detailed results of model evaluations are presented in Section \ref{Experiments/Result}.

\begin{figure}[!hpbt]
    \centering
    \includegraphics[width=\textwidth]{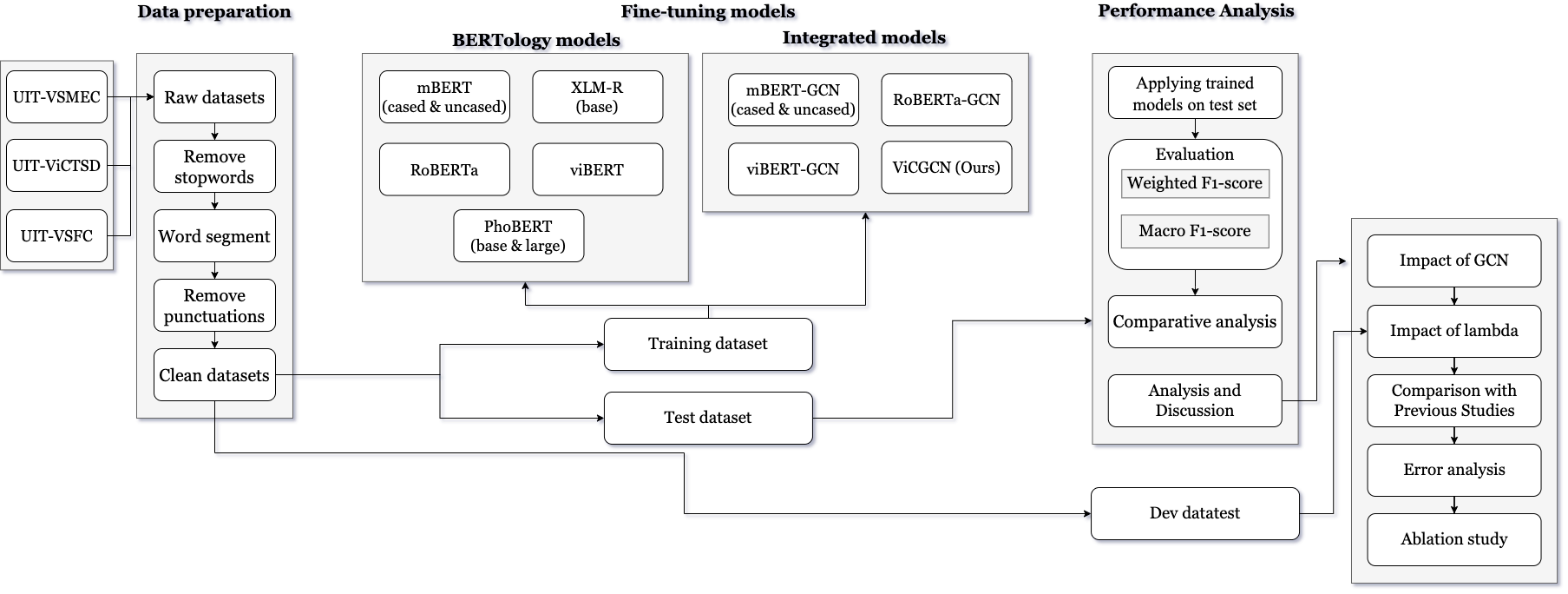}
    \caption{Overview of our experimental design.}
    \label{fig::Experiments/Procedure/Overview}
\end{figure}

To gain a deeper insight into our proposed model, we conduct a comprehensive analysis and discussion from various angles. This includes assessing the impact of graph convolutional networks (see Section \ref{imapactGCN}) and the influence of the lambda parameter (see Section \ref{impactlamda}). Additionally, we make comparisons with prior studies to accurately gauge the accomplishments of our research (see Section \ref{comparisonprestudies}). Furthermore, we select the model that exhibits the most outstanding performance to carry out an error analysis on the inaccuracies detected within our proposed model (see Section \ref{erroranalysis}). Towards the end of the experiment, we conduct an ablation study to investigate the effectiveness and contribution of our proposed approach, ViCGCN (see Section \ref{ablationstudy}). Figure \ref{fig::Experiments/Procedure/Overview} provides an overview of our methodology, encompassing data preparation, baseline fine-tuning, the proposed model, and performance analysis.

\subsection{Baseline Models} \label{Experiments/Baseline}
Contextualized language models have been extensively used in various natural language processing tasks, including text classification. Additionally, since PhoBERT and viBERT are monolingual models specifically designed for the Vietnamese language, comparing their performance with a widely used and established model like mBERT is essential. Furthermore, as GCN has been shown to effectively capture the context and relationships between words in a text, integrating it with a contextualized language model could improve its performance in text classification tasks. Because of the following reasons, we compare our ViCGCN model with baseline models.
\subsubsection{Contextualized Language Models}
\begin{itemize}
    \item \textbf{Multilingual BERT (mBERT)\footnote{\url{https://github.com/google-research/bert}}}: mBERT, introduced by \citet{devlin-etal-2019-bert}, is a BERT-based model with specific characteristics. It consists of 12 layers, 768 hidden units, 12 attention heads, and a total of 110 million parameters. Remarkably, mBERT is designed to support 104 distinct languages, and it has been trained on and can be applied to text in these 104 languages using a combination of masked language modeling (MLM) and next sentence prediction objectives. This training corpus includes content from Wikipedia\footnote{\url{https://www.wikipedia.org/}}. \textit{mBERT} consists of two versions cased\footnote{\url{https://huggingface.co/bert-base-multilingual-cased}} and uncased\footnote{\url{https://huggingface.co/bert-base-multilingual-uncased}}.
    \item \textbf{RoBERTa\footnote{\url{https://huggingface.co/roberta-base}}}: \citet{DBLP:journals/corr/abs-1907-11692} proposed RoBERTa. They utilize a dynamic masking technique during the training process, instructing the model to predict intentionally hidden segments of text within unannotated language samples. RoBERTa, implemented using the PyTorch framework, makes critical adjustments to BERT's essential hyperparameters.
    \item \textbf{XLM-RoBERTa (XLM-R)\footnote{\url{https://github.com/facebookresearch/XLM}}}: \citet{XLMR}  proposed XLM-R a masked language model based on the transformer architecture. This model stands out as a multilingual powerhouse, having been pre-trained on text from a staggering 100 languages. What makes XLM-R particularly impressive is the extensive and careful curation of over 2.5TB of data from CommonCrawl. Among its notable contributions are the improvements made for low-resource languages through specialized training and vocabulary expansion. Moreover, XLM-R boasts a more expansive shared vocabulary and a substantial increase in its overall model capacity, incorporating a whopping 550 million parameters. XLM-R includes \textit{base}\footnote{\url{https://huggingface.co/xlm-roberta-base}} and \textit{large}\footnote{\url{https://huggingface.co/xlm-roberta-large}} version.
    \item \textbf{PhoBERT\footnote{\url{https://huggingface.co/vinai/phobert-base}}}: \citet{nguyen-tuan-nguyen-2020-phobert} introduced a set of large-scale monolingual language models specifically designed for the Vietnamese language. Among these models, PhoBERT stands out as the state-of-the-art contextualized language model for Vietnamese. PhoBERT's architecture is built upon the RoBERTa model, but it has been optimized for training on a substantial Vietnamese corpus to effectively handle Vietnamese text. PhoBERT comes in two versions: \textit{base} and the \textit{large} versions.
    \item \textbf{viBERT\footnote{\url{https://huggingface.co/FPTAI/vibert-base-cased}}}: \citet{viBERT} introduced viBERT, a pre-trained language model for Vietnamese based on the BERT architecture. The architecture of viBERT is similar to that of mBERT, and it has been pre-trained on a large corpus of 10GB of uncompressed Vietnamese text. However, unlike mBERT, viBERT excludes insufficient vocabulary due to the inclusion of languages other than Vietnamese in the mBERT vocabulary.
    \item \textbf{vELECTRA\footnote{\url{https://huggingface.co/FPTAI/velectra-base-discriminator-cased}}}: \citet{viBERT}  unveiled vELECTRA, a pre-trained language model tailored for Vietnamese that adheres to the ELECTRA framework. vELECTRA shares a parallel architectural structure with ELECTRA and has undergone pretraining on an extensive corpus comprising 60GB of uncompressed Vietnamese text.
\end{itemize}

\subsubsection{Other Graph Neural Networks}
Bert-GCN was introduced by \citet{BertGCN}, presenting a novel approach that harnesses the benefits of extensive pretraining alongside transductive learning for the purpose of text classification. Bert-GCN achieves this by constructing a diverse graph over the dataset, where documents are represented as nodes, all leveraging the embedding power of BERT. Consequently, this research undertakes the implementation of various BERT variations, such as multilingual and Vietnamese monolingual models, in conjunction with GCN-combined models to assess their effectiveness in text classification for Vietnamese tasks. Additionally, when compared to mBERT-GCN, RoBERTa-GCN, viBERT-GCN, and vELECTRA-GCN, our proposed ViCGCN model offers valuable insights into the impact of integrating both monolingual and multilingual Contextualized Language Models with GCN on three standardized benchmark datasets. 

\subsection{Benchmark Datasets} \label{Experiments/Datasets}
\subsubsection{Benchmark Datasets} \label{Experiments/Datasets/Data}
To verify the efficiency of our proposed approach to text classification on Vietnamese social media, we conducted our experiments on three widely used Vietnamese social media corpora, including Vietnamese Social Media Emotion Corpus (UIT-VSMEC) that was made available by Ho et al. \citet{DBLP:journals/corr/abs-1911-09339}, Vietnamese Students' Feedback Corpus (UIT-VSFC) built by \citet{VSFC}, and Vietnamese Constructive and Toxic Speech Detection (UIT-ViCTSD) introduced by \citet{DBLP:journals/corr/abs-2103-10069}.

\begin{itemize}
    \item \textbf{UIT-VSMEC \citet{DBLP:journals/corr/abs-1911-09339}}: UIT-VSMEC consists of 6,927 sentences that have been annotated with emotions to tackle the challenge of identifying emotions in Vietnamese social media comments. This dataset encompasses seven emotion categories: Enjoyment, Disgust, Sadness, Anger, Fear, Surprise, and Other.
    \item \textbf{UIT-VSFC \citet{VSFC}}: UIT-VSFC comprises 16,000 sentences that have been investigated for two distinct purposes: one related to sentiment analysis and the other related to topic classification. The sentiment analysis task involves categorizing sentences into three classes: Positive, Negative, and Neutral. Meanwhile, the topic classification task involves assigning sentences to one of four categories: Lecturer, Curriculum, Facility, or Others.
    \item \textbf{UIT-ViCTSD \cite{DBLP:journals/corr/abs-2103-10069}}: UIT-ViCTSD consists of 10,000 human-annotated comments on ten domains. Each comment is categorized into two tasks: constructiveness and toxicity in Vietnamese social media, which are binary classifications. Two categories are used to denote feedback: constructive and non-constructive. Similarly, comments can be labeled as toxic or non-toxic to identify harmful behavior.
\end{itemize}

\subsubsection{Pre-processing techniques}
A few efficient pre-processing techniques for Vietnamese text in general and Vietnamese social media text in particular were presented \cite{nguyen2020exploiting, PhoBERT-CNN}. However, we only follow some simple preprocessed techniques according to the quality of the three benchmark datasets mentioned in Section \ref{Experiments/Datasets/Data} and more essential to prove the outperform and efficiency of our model ViCGCN on Vietnamese social media raw text. Firstly, we removed stopwords defined in Vietnamese stopwords dict\footnote{\url{https://github.com/stopwords/vietnamese-stopwords}}. We, then, segment sentences into words by applying Word Segmenter of VnCoreNLP\footnote{\url{https://github.com/vncorenlp/VnCoreNLP}} for all of the models. Finally, the Regex\footnote{\url{https://docs.python.org/3/library/re.html}} library in Python is used to remove all punctuations in three benchmark datasets.


The statistics of the pre-processed datasets are summarized in Table \ref{4/Dataset}.

    
\begin{table}[!ht]
\centering
\caption{Statistics and descriptions of tasks of each dataset in this study.}
\resizebox{\linewidth}{!}{%
\begin{tabular}{lrrrlr} 
\hline
\textbf{Dataset}            & \multicolumn{1}{l}{\textbf{Train}} & \multicolumn{1}{l}{\textbf{Dev}} & \multicolumn{1}{l}{\textbf{Test}} & \multicolumn{1}{c}{\textbf{Task}}         & \multicolumn{1}{l}{\textbf{Classes}}  \\ 
\hline
\multicolumn{6}{c}{\textit{Binary text classification}}                                                                                                                                                                     \\ 
\hline
\multirow{2}{*}{UIT-ViCTSD} & 7,000                              & 2,000                            & 1,000                             & Constructive speech detection             & 2                                     \\
                            & 7,000                              & 2,000                            & 1,000                             & Toxic speech detection                    & 2                                     \\ 
\hline
\multicolumn{6}{c}{\textit{Multi-class text classification}}                                                                                                                                                                \\ 
\hline
\multirow{2}{*}{UIT-VSMEC}  & 5,548                              & 686                              & 693                               & Emotion recognition (with Other label)    & 7                                     \\
                            & 4,527                              & 583                              & 589                               & Emotion recognition (without Other label) & 6                                     \\ 
\cline{1-1}
\multirow{2}{*}{UIT-VSFC}   & 11,426                             & 1,583                            & 3,166                             & Sentiment-based classification            & 3                                     \\
                            & 11,426                             & 1,583                            & 3,166                             & Topic-based classification                & 4                                     \\
\hline
\end{tabular}}
\label{4/Dataset}
\end{table}

\subsection{Evaluation Metric} \label{Experiments/Metrics}

This section outlines the performance evaluation criteria utilized in this research. In the realm of classification tasks, especially concerning the three datasets highlighted within this study, the conventional metric employed is the Average Macro F1-score (\%). However, given the significant class imbalances in the provided datasets, the most appropriate metric for this study is the Average Macro F1-score, which is derived as the harmonic mean of Precision and Recall. Furthermore, to facilitate comparisons with prior studies, we have also adopted relevant measures based on the metrics employed in those studies, such as the Average Weighted F1-score (\%) for both UIT-VSMEC and UIT-VSFC datasets.

 To compute the average macro F1-score, firstly, we calculate Precision and Recall by Equation (\ref{eq::Experiments/Metrics/Presision}) and Equation (\ref{eq::Experiments/Metrics/Recall}) respectively. Then, Equation (\ref{eq::Experiments/Metrics/F1-score}) is used to determine F1-score per class in the dataset. $tp$ are truly positive, $fp$ – false positive, $fn$ – false negative, and $tn$ – true negative counts, respectively.
\begin{equation}
    Precision = \frac{tp}{tp+fp} \label{eq::Experiments/Metrics/Presision}
\end{equation}
\begin{equation}
    Recall=\frac{tp}{tp+fn} \label{eq::Experiments/Metrics/Recall}
\end{equation}
\begin{equation}
    \textit{F1-score}=2\times\frac{Precision\times Recall}{Precision+Recall} \label{eq::Experiments/Metrics/F1-score}
\end{equation}

We compute the average macro F1-score (mF1) and weighted F1-score (wF1) after acquiring the F1 scores for all classes. Equation (\ref{eq::Experiments/Metrics/macro F1-score}) and Equation (\ref{eq::Experiments/Metrics/weighted F1-score}) present the macro F1-score and weighted F1-score, respectively, for multi-class classification for multi classes $C_{i}$, i $\in$ \{1, 2,... n\} (denoted for every class of the dataset). Where $\textit{F1-score}_{i}$ and $W_{i}$ are the \textit{F1-score} and weight of class \textit{i} of the dataset, respectively.

\begin{equation}
    \textit{mF1} = \frac{{\sum_{i=1}^{n} \textit{F1-score}_{i}}}{n} \label{eq::Experiments/Metrics/macro F1-score}
\end{equation}
\begin{equation}
    \textit{wF1} = \frac{\sum_{i=1}^{n} {\textit{F1-score}_{i} \times W_{i}}}{\sum_{i=1}^{n} W_{i}} \label{eq::Experiments/Metrics/weighted F1-score}
\end{equation}
\subsection{Experiment Configuration}
Section \ref{Experiments/Configuration/Baseline} and Section \ref{Experiments/Configuration/Proposed} provide our settings for both baselines and the proposed approach in detail.

\subsubsection{Basesline models' configuration} \label{Experiments/Configuration/Baseline}
We implemented many transfer learning models including mBERT both \textit{cased} and \textit{uncased}, $\text{RoBERTa}$, XLM-R, $\text{PhoBERT}_{base}$, $\text{PhoBERT}_{large}$, vELECTRA, and viBERT in this study. They run with their max sequence length of 256, batch size of 32, epoch of 10, and Adam optimizer \cite{https://doi.org/10.48550/arxiv.1412.6980} with a fixed learning rate of 2e-5.
 
\subsubsection{Our approach's configuration} \label{Experiments/Configuration/Proposed}
In our proposed approach, $\text{PhoBERT}_{base}$ is the output feature of the [CLS] token as the sentence node, followed by a feedforward layer to derive the final prediction. We use $\text{PhoBERT}_{base}$ pre-trained model from HuggingFace combined with a two-layer GCN to implement ViCGCN. We initialize Adam optimizer \cite{https://doi.org/10.48550/arxiv.1412.6980} with a fixed learning rate of 1e-3 and 1e-5 for the GCN and PhoBERT module, respectively. Moreover, PhoBERT runs with a 256 max sequence length.

\subsection{Experimental Results} \label{Experiments/Result}

\begin{table}[!ht]
\centering
\caption{F1-score performances of models on the test sets of various Vietnamese social media textual datasets. Improvement (1) and Improvement (2) denoted the improvement over BERTology models and the improvement over BERTology integrated with GCN models, respectively.}
\label{tab::Experiments/Result}
\resizebox{\linewidth}{!}{%
\begin{tabular}{c|cc|cc|cc|cc|cc|cc} 
\hline
\textbf{Datasets}        & \multicolumn{4}{c|}{\textbf{UIT-VSMEC}}                                               & \multicolumn{4}{c|}{\textbf{UIT-ViCTSD}}                                                & \multicolumn{4}{c}{\textbf{UIT-VSFC}}                                                     \\ 
\hline
\textbf{Tasks}           & \multicolumn{2}{c|}{\textbf{Seven labels}} & \multicolumn{2}{c|}{\textbf{Six labels}} & \multicolumn{2}{c|}{\textbf{Constructiveness}} & \multicolumn{2}{c|}{\textbf{Toxicity}} & \multicolumn{2}{c|}{\textbf{Sentiment-based}} & \multicolumn{2}{c}{\textbf{Topic-based}}  \\ 
\hline
                         & \textbf{wF1}   & \textbf{mF1}              & \textbf{wF1}   & \textbf{mF1}            & \textbf{wF1}   & \textbf{mF1}                  & \textbf{wF1}   & \textbf{mF1}          & \textbf{wF1}   & \textbf{mF1}                 & \textbf{wF1}   & \textbf{mF1}             \\ 
\hline
mBERT (\textit{cased)}   & 60.47          & 59.48                     & 65.02          & 62.65                   & 81.03          & 79.55                         & 88.32          & 65.63                 & 90.39          & 77.15                        & 87.32          & 77.93                    \\
mBERT (\textit{uncased)} & 60.17          & 59.18                     & 64.93          & 62.11                   & 80.89          & 79.47                         & 87.6           & 64.77                 & 89.95          & 77.8                         & 87.62          & 77.58                    \\
RoBERTa                  & 58.17          & 57.32                     & 63.32          & 59.97                   & 77.41          & 75.62                         & 85.85          & 59.71                 & 87.13          & 75.52                        & 86.77          & 75.30                    \\
XLM-R                    & 62.02          & 61.01                     & 68.19          & 63.70                   & 81.81          & 80.85                         & 89.92          & 73.09                 & 93.03          & 82.61                        & 89.67          & 79.25                    \\
PhoBERT \textit{base}    & 64.36          & 61.41                     & 69.02          & 64.12                   & 81.65          & 80.24                         & 89.58          & 72.12                 & 92.94          & 82.15                        & 88.29          & 78.54                    \\
PhoBERT \textit{large}   & 65.12          & 63.23                     & 71.13          & 65.12                   & 82.07          & 81.27                         & 90.12          & 73.32                 & 93.24          & 82.96                        & 88.72          & 79.12                    \\
vELECTRA                 & 63.58          & 61.38                     & 68.33          & 63.12                   & 82.41          & 80.82                         & 89.33          & 72.02                 & 91.89          & 82.01                        & 88.12          & 78.12                    \\
viBERT                   & 61.33          & 60.28                     & 68.48          & 62.09                   & 81.62          & 80.07                         & 89.14          & 71.87                 & 91.29          & 81.95                        & 88.22          & 78.35                    \\ 
\hline
mBERT-GCN (cased)        & 68.32          & 64.32                     & 69.32          & 66.18                   & 83.12          & 82.88                         & 90.32          & 69.42                 & 92.12          & 79.32                        & 88.32          & 79.42                    \\
mBERT-GCN (uncased)      & 67.98          & 64.11                     & 69.12          & 65.89                   & 82.32          & 82.01                         & 89.15          & 68.32                 & 91.01          & 79.02                        & 88.07          & 79.02                    \\
RoBERTa-GCN              & 66.17          & 62.12                     & 67.12          & 64.17                   & 81.33          & 80.96                         & 89.02          & 64.32                 & 90.12          & 78.42                        & 87.45          & 78.12                    \\
vELECTRA-GCN             & 69.42          & 65.44                     & 70.95          & 67.20                   & 84.62          & 84.62                         & 91.88          & 74.85                 & 93.56          & 83.12                        & 89.95          & 80.02                    \\
viBERT-GCN               & 69.32          & 65.12                     & 70.83          & 66.68                   & 84.32          & 83.12                         & 91.12          & 74.25                 & 93.12          & 82.47                        & 89.42          & 79.63                    \\ 
\hline
\textbf{ViCGCN (base)}   & \textbf{70.32} & \textbf{67.17}            & \textbf{71.02} & \textbf{67.48}          & \textbf{85.64} & \textbf{85.12}                & \textbf{92.22} & \textbf{75.32}        & \textbf{94.12} & \textbf{83.67}               & \textbf{90.12} & \textbf{80.11}           \\
\textbf{ViCGCN (large)}  & \textbf{71.33} & \textbf{67.82}            & \textbf{72.08} & \textbf{68.12}          & \textbf{86.12} & \textbf{85.88}                & \textbf{93.11} & \textbf{76.12}        & \textbf{94.83} & \textbf{84.23}               & \textbf{91.02} & \textbf{81.88}           \\ 
\hline
Improvement (1)          & $\uparrow$6.21 & $\uparrow$4.59            & $\uparrow$0.95 & $\uparrow$3.00          & $\uparrow$3.71 & $\uparrow$4.61                & $\uparrow$2.99 & $\uparrow$2.80        & $\uparrow$1.59 & $\uparrow$1.27               & $\uparrow$1.35 & $\uparrow$2.63           \\
Improvement (2)          & $\uparrow$1.91 & $\uparrow$2.38            & $\uparrow$1.13 & $\uparrow$0.92          & $\uparrow$1.50 & $\uparrow$1.26                & $\uparrow$1.23 & $\uparrow$1.27        & $\uparrow$1.27 & $\uparrow$1.11               & $\uparrow$1.07 & $\uparrow$1.86           \\
\hline
\end{tabular}}
\end{table}

To demonstrate the classification performance of our model ViCGCN, we compare it with other state-of-the-art and Integrated models as mentioned in Section \ref{Experiments/Baseline}. The F1-score results for both baseline and proposed models on the test sets of three Vietnamese social media text datasets are shown in Table \ref{tab::Experiments/Result} and we obtain the following observations.

Among BERTology models, RoBERTa and mBERT, including \textit{cased} and \textit{uncased}, have the most unfavorable performance of almost tasks of the three benchmark datasets. Moreover, the results show that monolingual models such as PhoBERT and viBERT perform better than other BERTology models. Additionally, through the execution of parallel computations for words, the problem of vanishing gradients is minimized, and PhoBERT archives the highest results in nearly all the tasks. However, in general, BERTology baseline models still find it hard to handle the complexity of social media: imbalanced and noisy data, which leads to poor performance compared to the integrated GCN model.
    
Our baseline integrated models can also benefit from graph structure by combining GCN as the final prediction module. Compared to BERTology baseline models, the performance boost from contextualized pre-trained language models with the GCN module is significant. Moreover, the multilingual and monolingual models integrated with GCN perform massively better than others. This explains the significance of incorporating both the Contextualized and GCN models into the integrated models can be attributed to their complementary nature in addressing the limitations of each other.

Compared to baseline models, our approach ViCGCN adopts large-scale, monolingual Vietnamese language model PhoBERT. Our integrated model ViCGCN obtains the ability to compute the new features of a node as the weighted average of itself and its second-order neighbors. In the context of imbalanced and noisy datasets, such as UIT-ViCTSD, the proposed ViCGCN model has demonstrated significant performance improvements compared to other baseline models, making it a promising approach for social media mining tasks. Moreover, Our proposed model demonstrated superior performance to the current state-of-the-art Vietnamese model, achieving improvements of 6.21\%, 4.61\%, and 2.63\% on three benchmark datasets. These results demonstrate the efficacy and validity of ViCGCN for Vietnamese text classification. As a result, our method achieves the best performance among all the tasks on three benchmark datasets in terms of UIT-VSMEC, UIT-ViCTSD, and UIT-VSFC, respectively.

\subsection{Analysis and Discussion}

\subsubsection{Impact of graph convolutional networks}
\label{imapactGCN}

Although we can implicitly infer the effectiveness of graph convolutional networks from Table \ref{tab::Experiments/Result}, we would like to discuss more the contribution of graph convolutional networks in contextualized language models. Table \ref{Result/Graph/VSMEC/table}, Table \ref{/Result/Graph/ViCTSD/table} and Table \ref{/Result/Graph/VSFC/table} display the comparisons between with and without GCN on three benchmark datasets as we can find that contextualized language model integrated with GCN outperformed all of the corresponding single models, respectively. As mentioned in Section \ref{Experiments/Result}, Contextualized Language Models have not performed well on three benchmark datasets. Integrating GCN with the BERTology model massively enhances the performance, which leads to improvements of up to 8.00\%, 7.99\%, 5.84\%, and 7.99\% of RoBERTa, viBERT, vELECTRA, and $\text{PhoBERT}_{base}$, respectively, on three benchmark datasets, UIT-VSMEC, UIT-ViCTSD, and UIT-VSFC, respectively. The average length of three datasets in UIT-VSMEC, UIT-ViCTSD, and UIT-VSFC is approximately 14. Additionally, the short sequence lengths can construct more dense graphs that provide richer contextual information, which may explain better performance by combining contextualized language models with GCN. This further demonstrates that Graph Convolutional Networks are essential in improving text classification performance.

\begin{table}[!ht]
\centering
\caption{Model performance on UIT-VSMEC.}
\label{Result/Graph/VSMEC/table}
\resizebox{\linewidth}{!}{%
\begin{tabular}{l|cc|cc} 
\hline
\textbf{Tasks}          & \multicolumn{2}{c|}{\textbf{Seven labels}}                        & \multicolumn{2}{c}{\textbf{Six labels}}                            \\ 
\hline
                        & \textbf{wF1}                    & \textbf{mF1}                    & \textbf{wF1}                    & \textbf{mF1}                     \\ 
\hline
mBERT (cased)           & 60.47                           & 59.48                           & 65.02                           & 62.65                            \\
mBERT-GCN (cased)       & 68.32 ($\uparrow$7.85)          & 64.32 ($\uparrow$4.84)          & 69.32 ($\uparrow$4.30)          & 66.18 ($\uparrow$3.53)           \\ 
\hline
mBERT (uncased)         & 60.17                           & 59.18                           & 64.93                           & 62.11                            \\
mBERT-GCN (uncased)     & 67.98 ($\uparrow$7.81)          & 64.11 ($\uparrow$4.93)          & 69.12 ($\uparrow$4.90)          & 65.89 ($\uparrow$3.78)           \\ 
\hline
RoBERTa                 & 58.17                           & 57.32                           & 63.32                           & 59.97                            \\
RoBERTa-GCN             & 66.17 ($\uparrow$8.00)          & 62.12 ($\uparrow$4.80)          & 67.12 ($\uparrow$3.80)          & 64.17 ($\uparrow$4.20)           \\ 
\hline
viBERT                  & 61.33                           & 60.28                           & 68.48                           & 62.09                            \\
viBERT-GCN              & 69.32 ($\uparrow$7.99)          & 78.37 ($\uparrow$4.84)          & 82.33 ($\uparrow$2.35)          & 81.98 ($\uparrow$4.59)           \\ 
\hline
vELECTRA                & 63.58                           & 61.38                           & 68.33                           & 63.12                            \\
vELETRA-GCN             & 69.42 ($\uparrow$5.84)          & 65.44 ($\uparrow$4.06)          & 70.95 ($\uparrow$2.62)          & 67.20 ($\uparrow$4.08)           \\ 
\hline
PhoBERT (base)          & 64.36                           & 61.41                           & 69.02                           & 64.12                            \\
\textbf{ViCGCN (base)}  & \textbf{69.32 ($\uparrow$7.99)} & \textbf{65.12 ($\uparrow$4.84)} & \textbf{70.83 ($\uparrow$2.35)} & \textbf{66.68 ($\uparrow$4.59)}  \\ 
\hline
PhoBERT (large)         & 65.12                           & 71.13                           & 63.23                           & 65.12                            \\
\textbf{ViCGCN (large)} & \textbf{71.33 ($\uparrow$6.21)} & \textbf{72.08 ($\uparrow$0.95)} & \textbf{67.82 ($\uparrow$4.59)} & \textbf{68.12 ($\uparrow$3.00)}  \\
\hline
\end{tabular}}
\end{table}

\begin{table}[H]
\centering
\caption{Model performance on UIT-ViCTSD.}
\label{/Result/Graph/ViCTSD/table}
\resizebox{\linewidth}{!}{%
\begin{tabular}{l|cc|cc} 
\hline
\textbf{Tasks}            & \multicolumn{2}{c|}{\textbf{Constructiveness}}                    & \multicolumn{2}{c}{\textbf{Toxicity}}                              \\ 
\hline
                          & \textbf{wF1}                    & \textbf{mF1}                    & \textbf{wF1}                    & \textbf{mF1}                     \\ 
\hline
mBERT (mBERT cased)       & 81.03                           & 79.55                           & 88.32                           & 65.63                            \\
mBERT-GCN (mBERT cased)   & 83.12 ($\uparrow$2.09)          & 82.88 ($\uparrow$3.33)          & 90.32 ($\uparrow$2.00)          & 69.42 ($\uparrow$3.79)           \\ 
\hline
mBERT (uncased)           & 80.89                           & 79.47                           & 87.60                           & 64.77                            \\
mBERT-GCN (mBERT uncased) & 82.32 ($\uparrow$1.43)          & 82.01 ($\uparrow$2.54)          & 89.15 ($\uparrow$1.55)          & 68.32 ($\uparrow$3.55)           \\ 
\hline
RoBERTa                   & 77.41                           & 75.62                           & 85.85                           & 59.71                            \\
RoBERTa-GCN               & 81.33 ($\uparrow$3.92)          & 80.96 ($\uparrow$5.34)          & 89.02 ($\uparrow$3.17)          & 64.32 ($\uparrow$4.61)           \\ 
\hline
viBERT                    & 81.62                           & 80.07                           & 89.14                           & 71.87                            \\
viBERT-GCN                & 84.32 ($\uparrow$2.70)          & 83.12 ($\uparrow$3.05)          & 91.12 ($\uparrow$1.98)          & 74.25 ($\uparrow$2.38)           \\ 
\hline
vELECTRA                  & 82.41                           & 80.82                           & 89.33                           & 72.02                            \\
vELETRA-GCN               & 84.62 ($\uparrow$2.21)          & 84.62 ($\uparrow$3.80)          & 91.88 ($\uparrow$2.55)          & 74.85 ($\uparrow$2.83)           \\ 
\hline
PhoBERT (base)            & 81.65                           & 80.24                           & 89.58                           & 72.12                            \\
\textbf{ViCGCN (base)}    & \textbf{85.64 ($\uparrow$3.99)} & \textbf{85.12 ($\uparrow$4.88)} & \textbf{92.22 ($\uparrow$2.64)} & \textbf{75.32 ($\uparrow$3.20)}  \\ 
\hline
PhoBERT large             & 82.07                           & 90.12                           & 81.27                           & 73.32                            \\
\textbf{ViCGCN (large)}   & \textbf{86.12 ($\uparrow$4.05)} & \textbf{93.11 ($\uparrow$2.99)} & \textbf{85.88 ($\uparrow$4.61)} & \textbf{76.12 ($\uparrow$2.80)}  \\
\hline
\end{tabular}}
\end{table}

\begin{table}[!ht]
\centering
\caption{Model performance on UIT-VSFC.}
\label{/Result/Graph/VSFC/table}
\resizebox{\linewidth}{!}{%
\begin{tabular}{l|cc|cc} 
\hline
\textbf{Tasks}            & \multicolumn{2}{c|}{\textbf{Sentiment-based}}                                                                                          & \multicolumn{2}{c}{\textbf{Topic-based}}                                             \\ 
\hline
                          & \textbf{wF1}                                                                                & \textbf{mF1}                             & \textbf{wF1}                             & \textbf{mF1}                              \\ 
\hline
mBERT (cased)             & 90.39                                                                                       & 77.15                                    & 87.32                                    & 77.93                                     \\
mBERT-GCN (cased)         & 92.12 ($\uparrow$1.73)                                                                      & 79.32 ($\uparrow$2.17)                   & 88.32 ($\uparrow$1.00)                   & 79.42 ($\uparrow$1.49)                    \\
mBERT (uncased)           & 89.95                                                                                       & 77.80                                    & 87.62                                    & 77.58                                     \\
mBERT-GCN (mBERT uncased) & 91.01 ($\uparrow$1.06)                                                                      & 79.02 ($\uparrow$1.22)                   & 88.07 ($\uparrow$0.45)                   & 79.02 ($\uparrow$1.44)                    \\ 
\hline
RoBERTa                   & 87.13                                                                                       & 75.52                                    & 86.77                                    & 75.30                                     \\
RoBERTa-GCN               & 90.12 ($\uparrow$2.99)                                                                      & 78.42 ($\uparrow$2.90)                   & 87.45 ($\uparrow$0.68)                   & 78.12 ($\uparrow$2.82)                    \\ 
\hline
viBERT                    & 91.29                                                                                       & 81.95                                    & 88.22                                    & 78.35                                     \\
viBERT-GCN                & 93.12 ($\uparrow$1.83)                                                                      & 82.47 ($\uparrow$0.52)                   & 89.42 ($\uparrow$1.20)                   & 79.63 ($\uparrow$1.28)                    \\ 
\hline
vELECTRA                  & 91.89                                                                                       & 82.01                                    & 88.12                                    & 78.12                                     \\
vELETRA-GCN               & 93.56 ($\uparrow$1.67)                                                                      & 83.12 $(\uparrow$1.11)                   & 89.95 ($\uparrow$1.83)                   & 80.02 ($\uparrow$1.90)                    \\ 
\hline
PhoBERT (base)            & 92.94                                                                                       & 82.15                                    & 88.29                                    & 78.54                                     \\
\textbf{ViCGCN (base)}    & \begin{tabular}[c]{@{}c@{}}\textbf{94.12~}($\uparrow$\textbf{}\textbf{1.18)}\end{tabular} & \textbf{83.67~}($\uparrow$\textbf{1.52)} & \textbf{90.12~}($\uparrow$\textbf{1.83)} & \textbf{80.11~}($\uparrow$\textbf{1.57)}  \\ 
\hline
PhoBERT (large)           & 93.24                                                                                       & 88.72                                    & 82.96                                    & 79.12                                     \\
\textbf{ViCGCN (large)}   & \textbf{94.83~}($\uparrow$\textbf{1.59)}                                                    & \textbf{91.02~}($\uparrow$\textbf{2.3)}  & \textbf{84.23~}($\uparrow$\textbf{1.27)} & \textbf{81.88~}($\uparrow$\textbf{2.76)}  \\
\hline
\end{tabular}}
\end{table}

\subsubsection{Impact of lambda ($\lambda$)}
\label{impactlamda}

\begin{figure}[!hpt]
    \centering

    \subfigure[Seven labels task]{\includegraphics[width=0.49\textwidth]{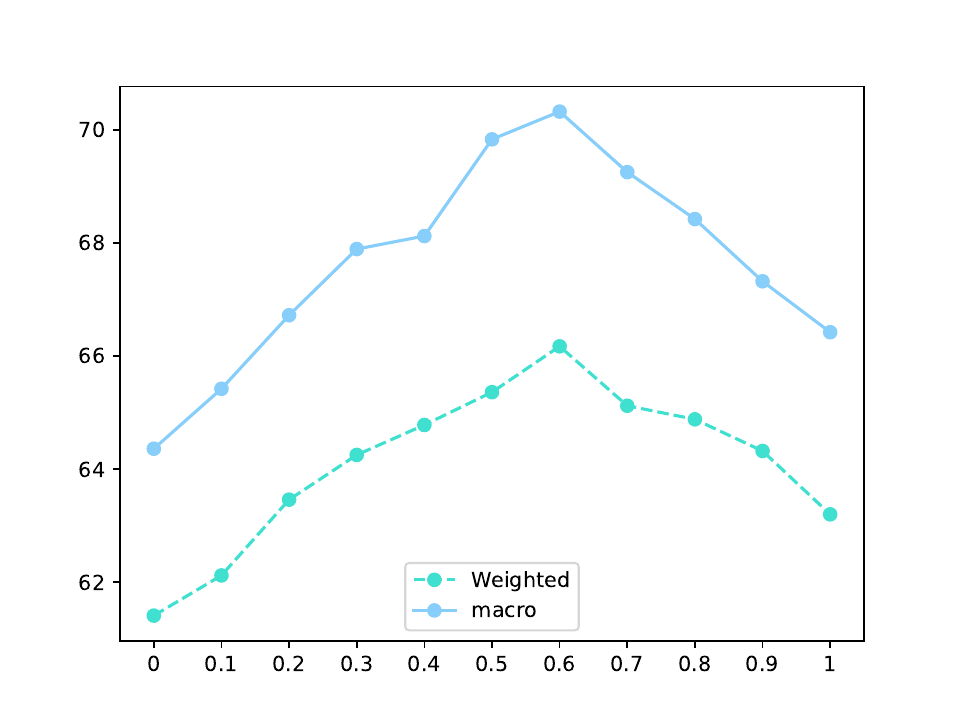}}
    \subfigure[Six labels task]{\includegraphics[width=0.49\textwidth]{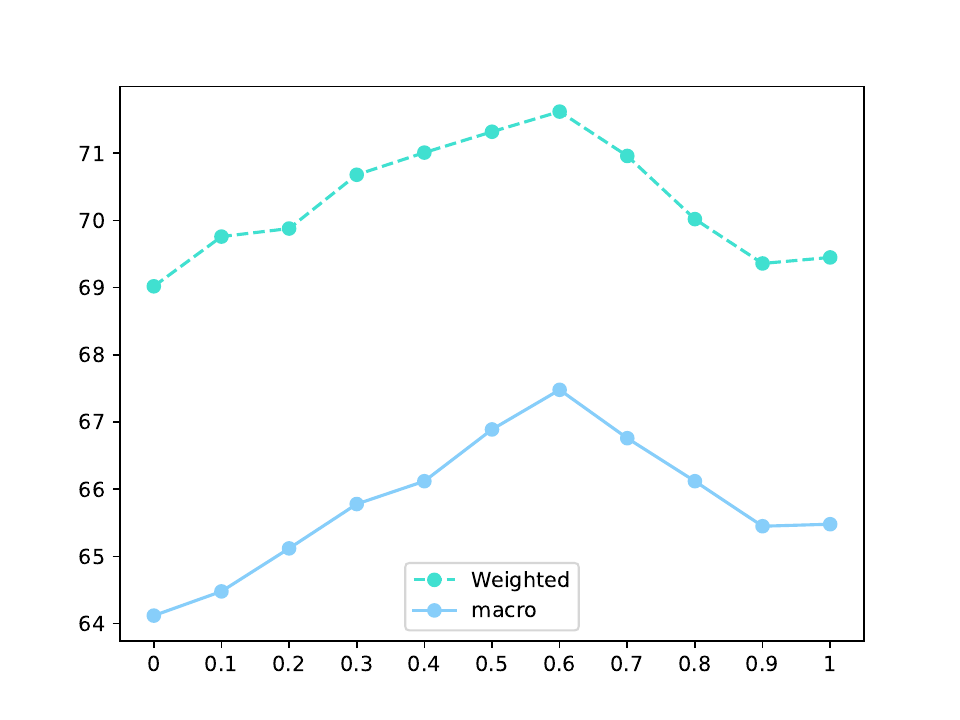}}

     \subfigure[Constructiveness task]{\includegraphics[width=0.49\textwidth]{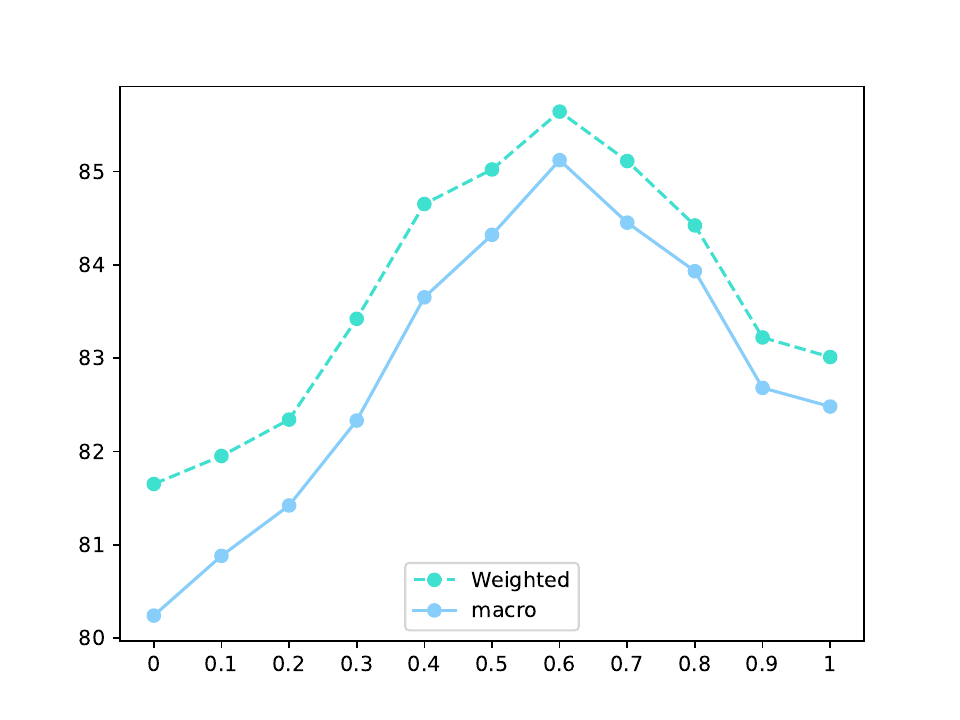}}
    \subfigure[Toxicity task]{\includegraphics[width=0.49\textwidth]{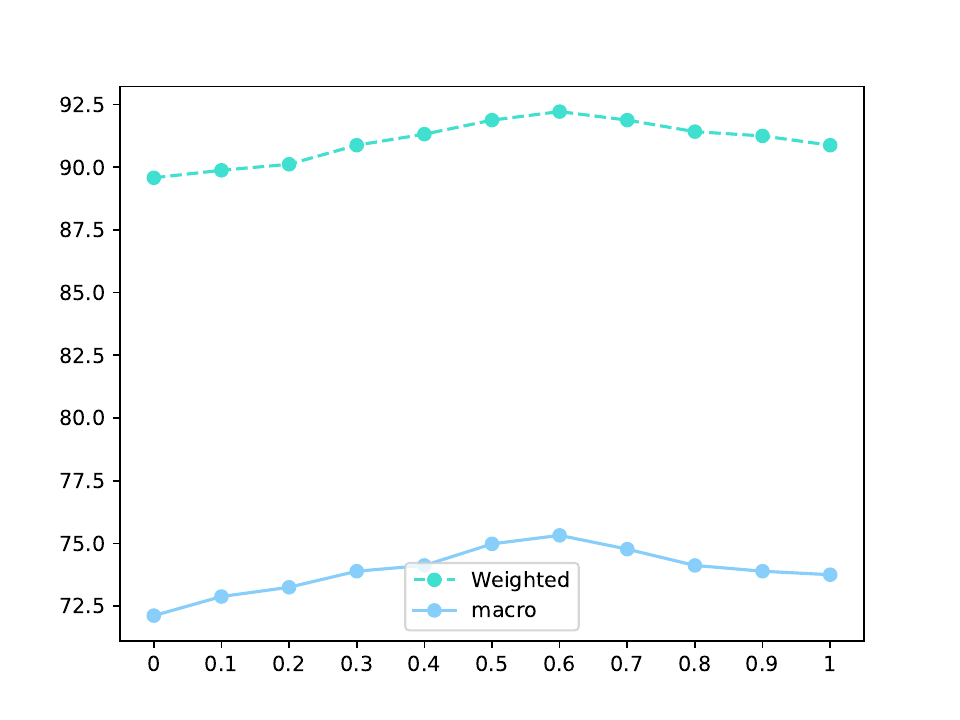}}
    
    \subfigure[Sentiment-based task]{\includegraphics[width=0.49\textwidth]{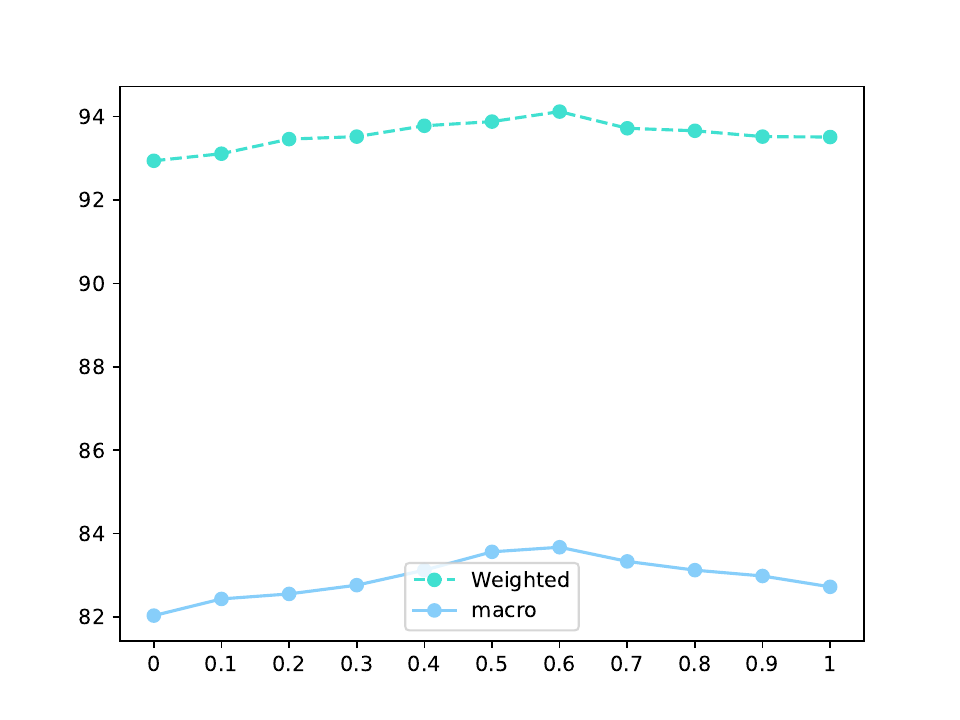}}
    \subfigure[Topic-based task]{\includegraphics[width=0.49\textwidth]{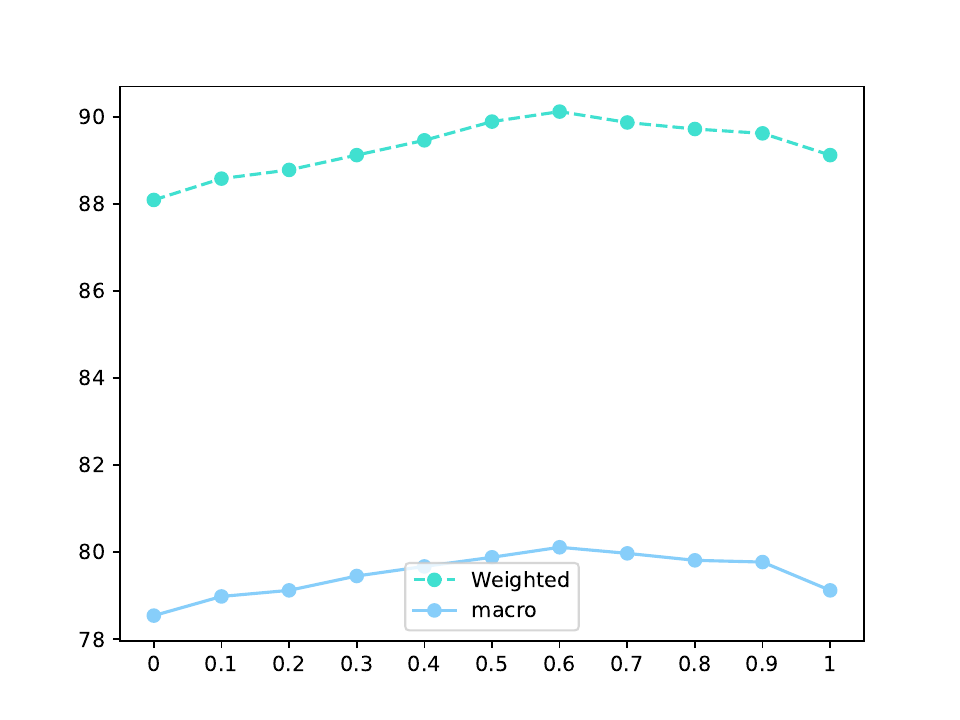}}
    
    \caption{F1-score of ViCGCN when varying $\lambda$ on the dev set.}
    
    \label{fig::Experiments/Lamda/UIT-VSFC}
\end{figure}

According to Equation \ref{equa::lambda}, the hyperparameter $\lambda$ controls the trade-off between two objectives, ViCGCN and PhoBERT, respectively. The optimal value of $\lambda$ may vary depending on the task. Therefore, extensive experiments on the dev set were conducted to determine the optimal value of $\lambda$. Figure \ref{fig::Experiments/Lamda/UIT-VSFC} shows the performances of ViCGCN on three benchmark datasets in terms of UIT-VSMEC, UIT-ViCTSD, and UIT-VSFC with different $\lambda$. On all three benchmark datasets, the F1-score is consistently higher with a more enormous $\lambda$ value. Moreover, taking only ViCGCN ($\lambda = 1$) as the final training objective consistently achieves a better performance than considering only PhoBERT ($\lambda = 0$). Setting $\lambda$ to a value from 0.6 to 0.8 is more desirable and can make the model reach its best when $\lambda = 0.6$ on all datasets. These observations indicate that the linear interpolation of the prediction from ViCGCN and the prediction from PhoBERT with higher ViCGCN weight can improve the Vietnamese social media text classification performance. On the other hand, the PhoBERT module is also indispensable.

\subsubsection{Comparison with Previous Studies}
\label{comparisonprestudies}
We conducted a number of surveys to evaluate how well our suggested technique performed in comparison to earlier studies. On the UIT-VSMEC, UIT-ViCTSD, and UIT-VSFC datasets, our method fared better than in any prior research. To provide for fair comparisons, similar evaluation metrics from earlier studies are employed. For all datasets used in this study, we use the average macro F1-score (\%) and average weighted F1-score (\%). 

Our integrated model ViCGCN outperformed the best results of each previous study on the VSMEC dataset by achieving 80.24\% weighted F1-score and 80.96\% macro F1-score on task Seven labels, which improves by 10.18\% and 13.93\% compared to the best previous study. Additionally, our model obtains 84.91\% weighted F1-score on the Six labels task as shown in Table \ref{tab::Experiments/Comparison/VSMEC}, increased by 13.92\% in comparison to the highest previous ones. Furthermore, Table \ref{tab::Experiments/Comparison/ViCTSD} deputed that ViCGCN achieves the best results, with a macro F1-score of 85.81\% for UIT-ViCTSD Constructiveness task, and 76.29\% macro F1-score for UIT-ViCSTD Toxicity task,  increased by 16.89\%. By obtaining 88.80\% macro F1-score and 94.81\% weighted F1-score, 89.61\% macro F1-score, and 93.81\% weighted F1-score on task Sentiment-based and Topic-based, respectively, our integrated model ViCGCN surpassed every previous study's top result on the UIT-VSFC dataset as describes in Table \ref{tab::Experiments/Comparison/VSFC}. In addition, our proposed approach reached new state-of-the-art performances on three Vietnamese benchmark social media datasets, UIT-VSMEC, UIT-ViCTSD, and UIT-VSFC, respectively. As a result, the proposed approach ViCGCN is significantly suitable and efficient for dealing with Vietnamese text in general and Vietnamese social media text classification tasks in particular.

\begin{table}[!hpt]
\centering
\caption{The comparison with previous studies on UIT-VSMEC.} \label{tab::Experiments/Comparison/VSMEC}
\resizebox{\linewidth}{!}{
\begin{tabular}{l|cc|cc} 
\hline
\textbf{Tasks}                                  & \multicolumn{2}{c|}{\textbf{Seven labels }} & \multicolumn{2}{c}{\textbf{Six labels }}  \\ 
\hline
                                                & \textbf{wF1}   & \textbf{mF1}               & \textbf{wF1}   & \textbf{mF1}             \\ 
\hline
CNN + Word2Vec                                  & 59.74          & -                          & 66.34          & -                        \\
MLR + TF-IDF Vectorizer + Key-clause extraction & 64.40          & -                          & -              & -                        \\
GRU + CNN + BiLSTM + LSTM                       & 65.79          & -                          & 70.99          & -                        \\
PhoBERT                                         & -              & 65.44                      & -              & -                        \\
XLM-R + VnEmolex                                & 70.06          & 67.03                      & -              & -                        \\ 
\hline
\textbf{ViCGCN (base)}                          & \textbf{70.32} & \textbf{67.17}             & \textbf{71.02} & \textbf{67.48}           \\
\textbf{ViCGCN (large)}                         & \textbf{71.33} & \textbf{67.82}             & \textbf{72.08} & \textbf{68.12}           \\
\hline
\end{tabular}}
\end{table}

\begin{table}[!ht] 
\centering
\caption{The comparison with previous studies on UIT-ViCTSD.}
\label{tab::Experiments/Comparison/ViCTSD}
\begin{tabular}{l|cc|cc} 
\hline
\textbf{Tasks}          & \multicolumn{2}{c|}{\textbf{Constructiveness }} & \multicolumn{2}{c}{\textbf{Toxicity }}  \\ 
\hline
                        & \textbf{wF1}   & \textbf{mF1}                   & \textbf{wF1}   & \textbf{mF1}           \\ 
\hline
PhoBERT                 & -              & 78.59                          & -              & 59.40                  \\
viBERT4news             & -              & 84.15                          & -              & -                      \\ 
\hline
\textbf{ViCGCN (base)}  & \textbf{85.64} & \textbf{85.12}                 & \textbf{92.22} & \textbf{75.32}         \\
\textbf{ViCGCN (large)} & \textbf{86.12} & \textbf{85.88}                 & \textbf{93.11} & \textbf{76.12}         \\
\hline
\end{tabular}
\end{table}

\begin{table}[!ht] 
\centering
\caption{The comparison with previous studies on UIT-VSFC.}
\label{tab::Experiments/Comparison/VSFC}
\resizebox{\linewidth}{!}{
\begin{tabular}{l|cc|cc} 
\hline
\textbf{Tasks}             & \multicolumn{2}{c|}{\textbf{Sentiment-based }} & \multicolumn{2}{c}{\textbf{Topic-based }}  \\ 
\hline
                           & \textbf{wF1}   & \textbf{mF1}                  & \textbf{wF1}   & \textbf{mF1}              \\ 
\hline
Maximum Entropy            & 87.64          & -                             & 84.03          & -                         \\
BiLSTM +Word2Vec~          & 92.03          & -                             & 89.62          & -                         \\
LD + SVM ()                & 92.20          & -                             & -              & -                         \\
BERT + CNN + BiLSTM + LSTM & 92.79          & -                             & 89.38          & -                         \\
BERT + CNN + BiLSTM        & 92.13          & -                             & 89.70          & -                         \\
XLM-R + VnEmoLex           & 93.97          & 83.40                         & -              & -                         \\ 
\hline
\textbf{ViCGCN (base)}     & \textbf{94.12} & \textbf{83.67}                & \textbf{90.12} & \textbf{80.11}            \\
\textbf{ViCGCN (large)}    & \textbf{94.83} & \textbf{84.23}                & \textbf{91.02} & \textbf{81.88}            \\
\hline
\end{tabular}}
\end{table}

\subsubsection{Errors Analysis}
\label{erroranalysis}

\begin{figure}[!ht]
    \centering
    \subfigure[Seven labels task]{\includegraphics[width=0.49\textwidth]{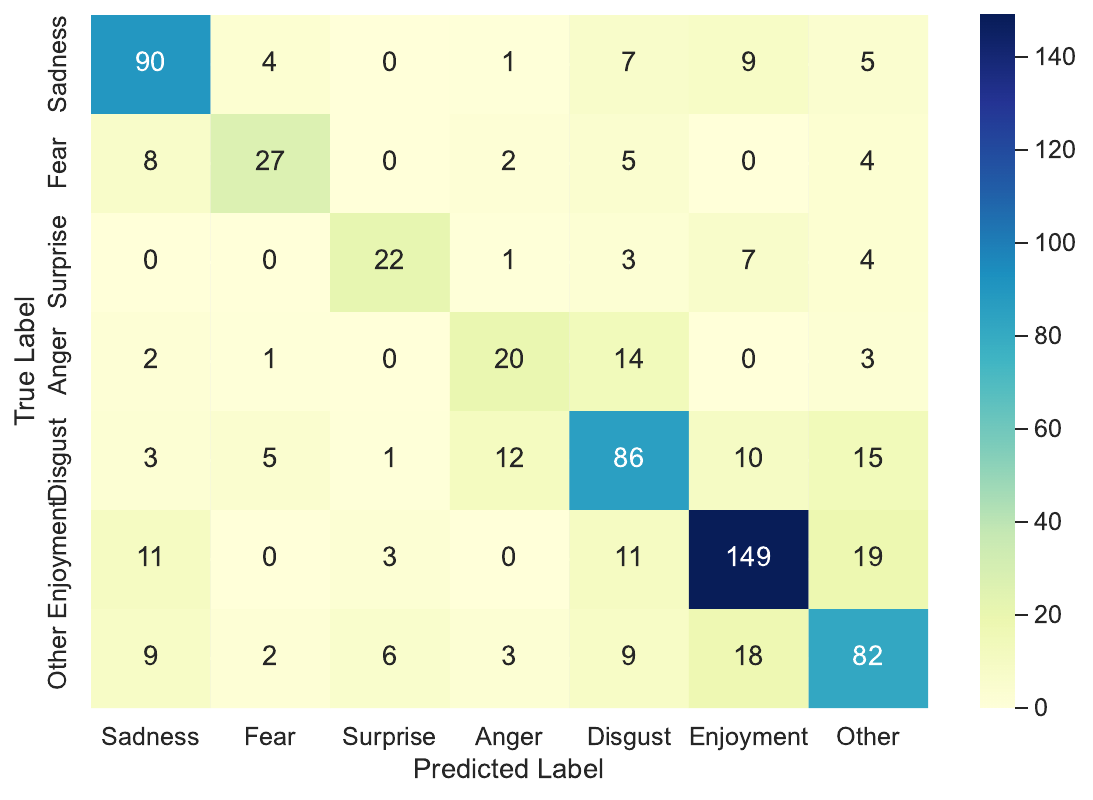}}
    \subfigure[Six labels task]{\includegraphics[width=0.49\textwidth]{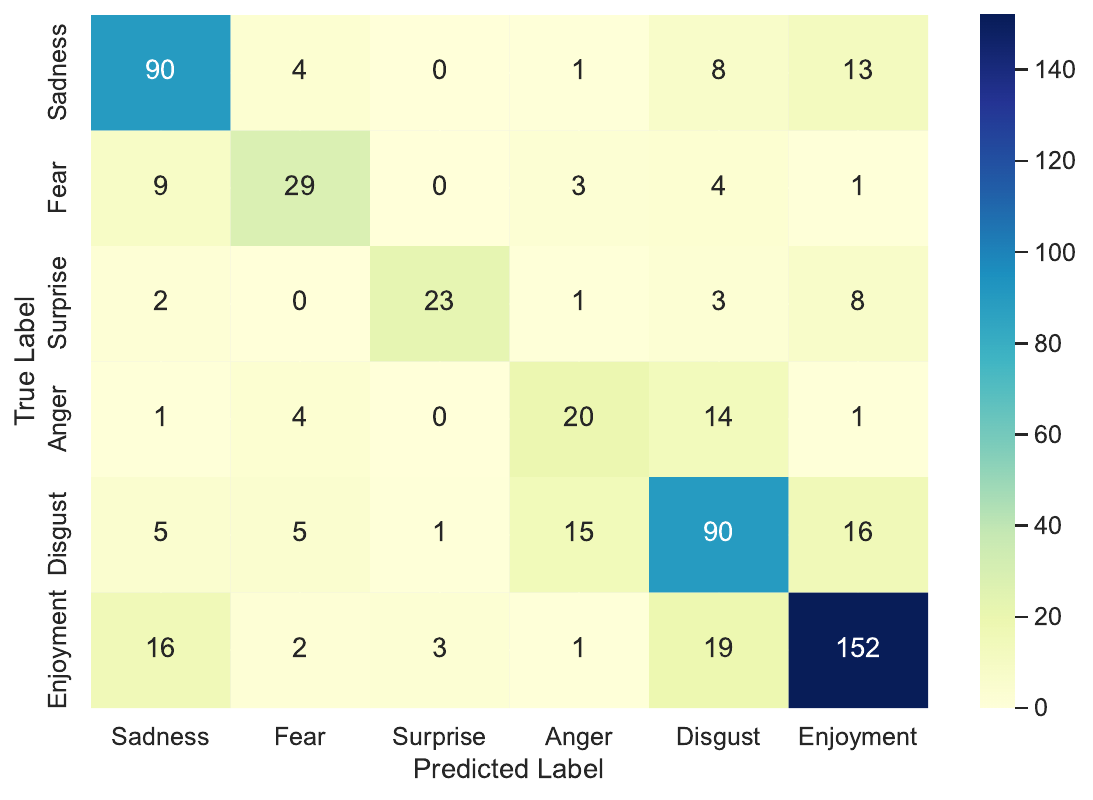}}
    \caption{Error analysis of our proposed approach for UIT-VSMEC dataset.}
    \label{fig::Experiments/CfMatrix/VSMEC}
\end{figure}

\begin{figure}[!ht]
    \centering
    \subfigure[Constructiveness task]{\includegraphics[width=0.49\textwidth]{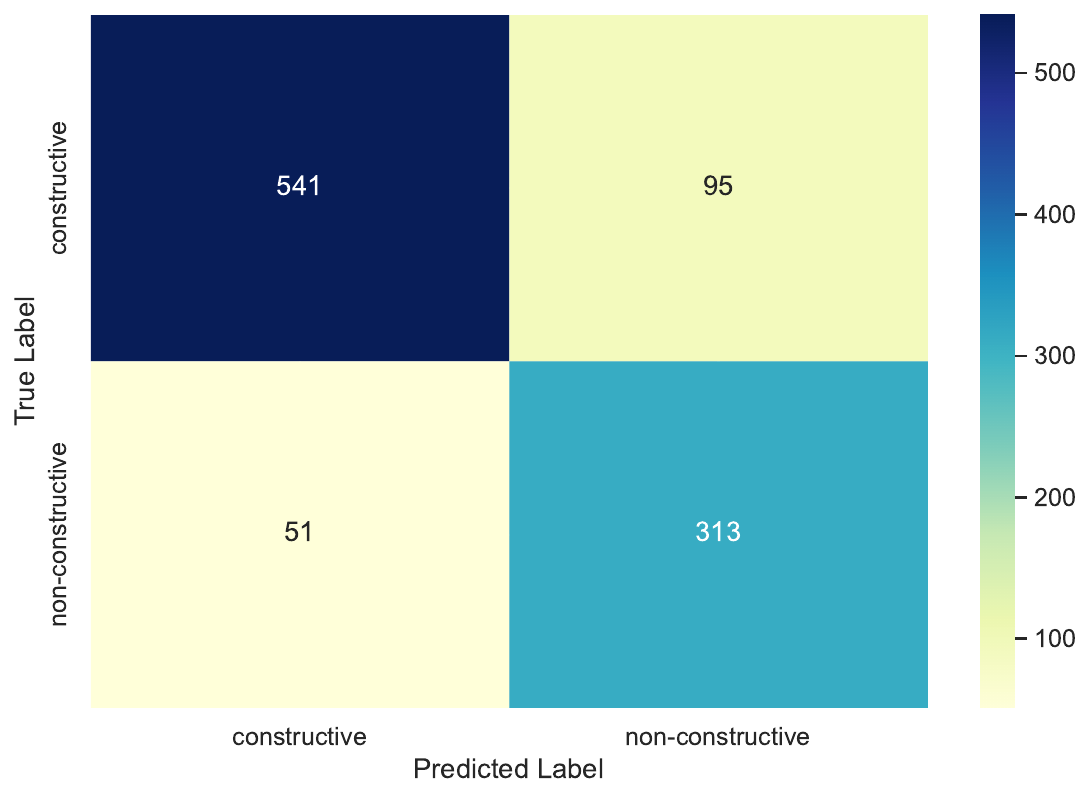}}
    \subfigure[Toxicity task]{\includegraphics[width=0.49\textwidth]{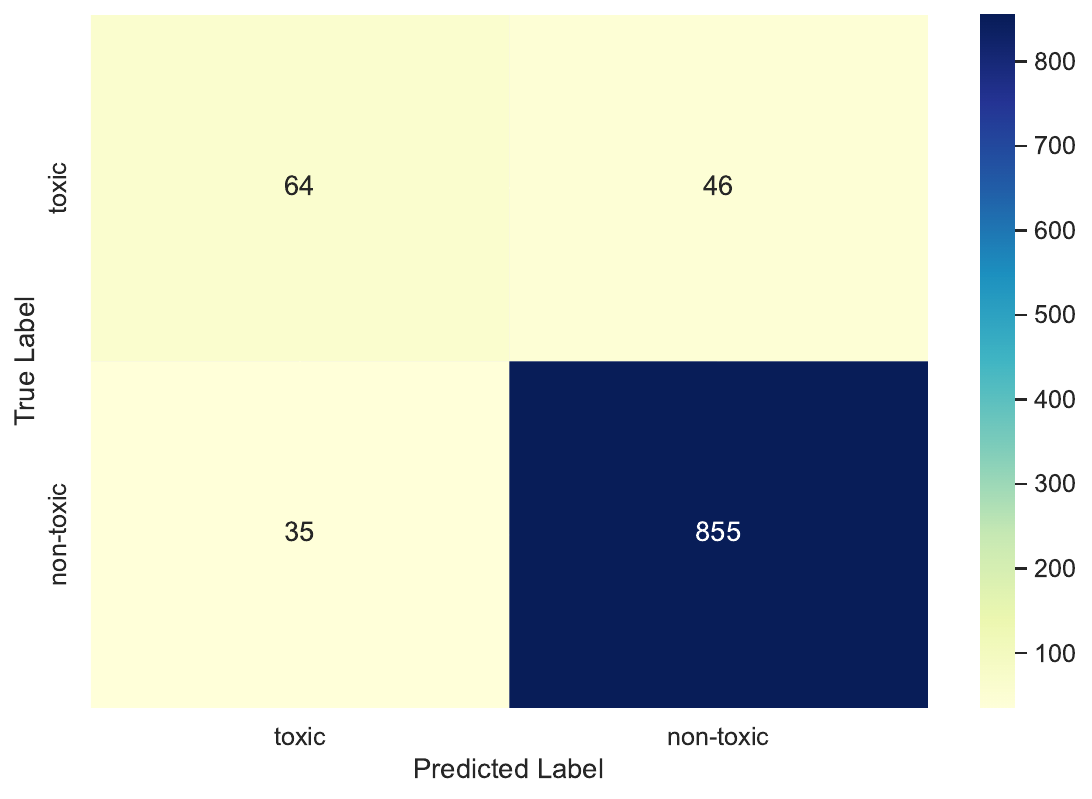}}
    \caption{Error analysis of our proposed approach for UIT-ViCTSD dataset.}
    \label{fig::Experiments/CfMatrix/ViCTSD}
\end{figure}

\begin{figure}[!ht]
    \centering
    \subfigure[Sentiment-based task]{\includegraphics[width=0.49\textwidth]{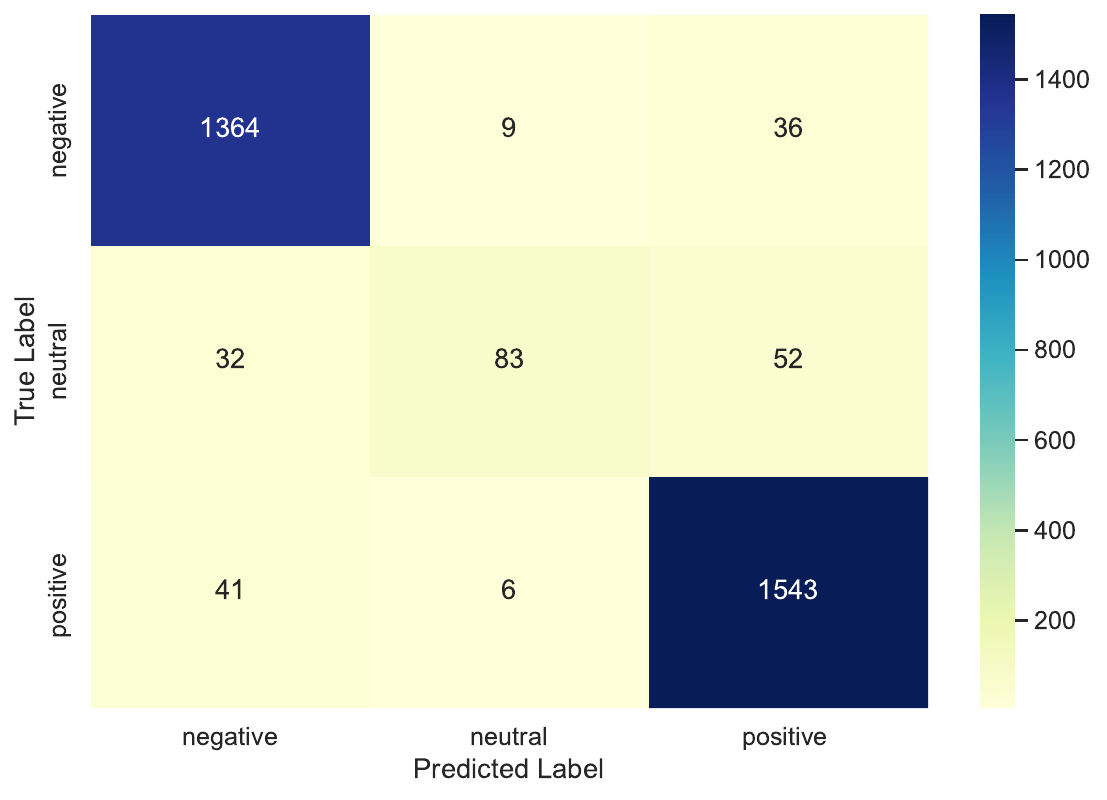}}
    \subfigure[Topic-based task]{\includegraphics[width=0.49\textwidth]{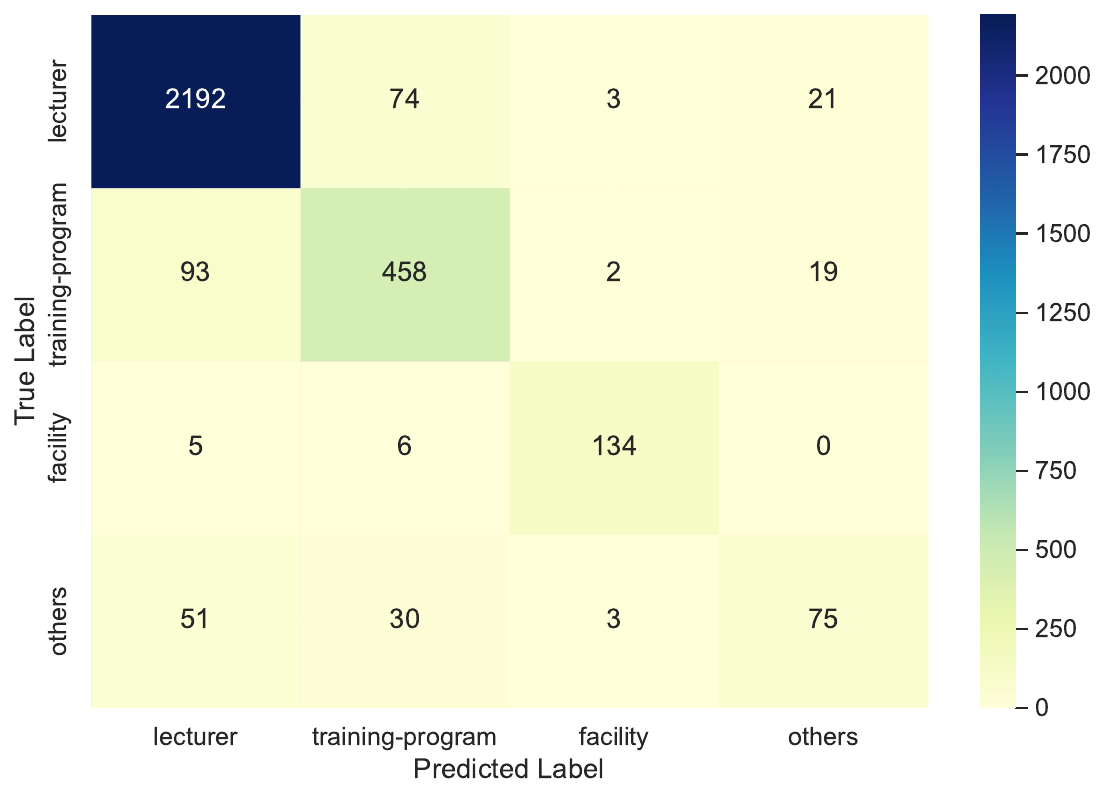}}
    \caption{Error analysis of our proposed approach for UIT-VSFC dataset.}
    \label{fig::Experiments/CfMatrix/VSFC}
\end{figure}

We utilize the error analysis of ViCGCN, our top-performing model, to analyze the errors observed in our proposed model. Figure\ref{fig::Experiments/CfMatrix/VSMEC}, Figure \ref{fig::Experiments/CfMatrix/ViCTSD} and Figure \ref{fig::Experiments/CfMatrix/VSFC}, respectively, show the confusion matrices for our best model's predictions on the test set for UIT-VSMEC, UIT-ViCTSD, and UIT-VSFC.


As described in Section \ref{Proposed model}, by incorporating contextualized language models such as BERT into GCN, ViCGCN can better capture the context and meaning of words and phrases, which can lead to more accurate identification of critical nodes. However, ViCGCN may not be able to explain why those nodes are essential or why specific nodes were not influential in the decision-making process. This can make it difficult for researchers to address specific issues in our proposed approach. Table \ref{fig:erroranalysissampleonViCTSD}, Table \ref{fig:erroranalysissampleonVSMEC}, and Table \ref{fig:erroranalysissampleonVSFC} contain a few illustrations of prediction errors. The results show that misclassifications were primarily due to the use of sarcasm, irony, and figurative language in social media comments. Furthermore, some misclassifications were due to the presence of multiple topics in a single comment, making it challenging to identify the primary intention. Additionally, ambiguity in identifying the labels of the datasets also leads to misclassifying of our proposed approach ViCGCN.


\begin{table}[H]
\centering
\caption{Several examples of classification error on UIT-VSMEC dataset.}\label{fig:erroranalysissampleonVSMEC}
\resizebox{\linewidth}{!}{%
\begin{tblr}{
  row{1} = {c},
  cell{2}{2} = {c},
  cell{2}{3} = {c},
  cell{3}{2} = {c},
  cell{3}{3} = {c},
  hline{1-2,4} = {-}{},
}
\textbf{Comment}                                                       & \textbf{True Label} & \textbf{Predicted Label} \\
{mấy ai được như vậy\\(\textbf{English:} not many people can do that)} & other               & surprise               \\
{kinh khủng thật\\(\textbf{English:} it's terrible)}                   & fear                & sadness                
\end{tblr}}
\end{table}


\begin{table}[!ht]
\centering
\caption{Several examples of classification error on UIT-ViCTSD dataset.}\label{fig:erroranalysissampleonViCTSD}
\resizebox{\linewidth}{!}{%
\begin{tblr}{
  row{1} = {c},
  cell{2}{2} = {c},
  cell{2}{3} = {c},
  cell{3}{2} = {c},
  cell{3}{3} = {c},
  hline{1-2,4} = {-}{},
}
\textbf{Comment}                                                                                                                                           & \textbf{True Label} & \textbf{Predicted Label} \\
{Người ăn không hết kẻ lần không ra\\(\textbf{English:} This man has much to eat but that \\
man finds no small piece.)}                                       & non\_constructive   & constructive           \\
{người trẻ còn sức khoẻ k lo làm ăn đi ăn trộm\\(\textbf{English:} Young people who are still healthy \\ don't worry about doing business but go to steal)} & non\_toxic          & toxic                  
\end{tblr}}
\end{table}


\begin{table}[!ht]
\centering
\caption{Several examples of classification error on UIT-VSFC dataset.}\label{fig:erroranalysissampleonVSFC}
\resizebox{\linewidth}{!}{%
\begin{tblr}{
  row{1} = {c},
  cell{2}{2} = {c},
  cell{2}{3} = {c},
  cell{3}{2} = {c},
  cell{3}{3} = {c},
  hline{1-2,4} = {-}{},
}
\textbf{Comment}                                                                                                                                          & \textbf{True Label} & \textbf{Predicted Label} \\
{ví dụ phù hợp với nội dung kiến thức , hướng dẫn chi tiết\\(\textbf{English:}~Examples are consistent with content knowledge, \\ detailed instructions)} & neural          & positive           \\
{đảm bảo chất lượng tốt\\(\textbf{English:}~Good quality guarantee)}                                                                                               & others          & facility topic     
\end{tblr}}
\end{table}

\subsubsection{Ablation Study}
\label{ablationstudy}

\begin{table}[H]
\centering
\caption{Ablation test on our proposed approach. w/o GCN and w/o PhoBERT denoted the result of the ablation GCN and the result of the ablation PhoBERT, respectively}
\label{tab::Ablation}
\resizebox{\linewidth}{!}{%
\begin{tabular}{l|cc|cc|cc|cc|cc|cc} 
\hline
\textbf{Datasets} & \multicolumn{4}{c|}{\textbf{VSMEC}}                                                                          & \multicolumn{4}{c|}{\textbf{ViCTSD}}                                                                         & \multicolumn{4}{c}{\textbf{VSFC}}                                                                \\ 
\hline
\textbf{Tasks}    & \multicolumn{2}{c|}{\textbf{Seven labels}}            & \multicolumn{2}{c|}{\textbf{Six labels}}             & \multicolumn{2}{c|}{\textbf{Constructiveness}}       & \multicolumn{2}{c|}{\textbf{Toxicity}}                & \multicolumn{2}{c|}{\textbf{Sentiment-based}}        & \multicolumn{2}{c}{\textbf{Topic-based}}  \\ 
\hline
                  & \multicolumn{1}{c|}{\textbf{wF1}} & \textbf{mF1}      & \multicolumn{1}{c|}{\textbf{wF1}} & \textbf{mF1}     & \multicolumn{1}{c|}{\textbf{wF1}} & \textbf{mF1}     & \multicolumn{1}{c|}{\textbf{wF1}} & \textbf{mF1}      & \multicolumn{1}{c|}{\textbf{wF1}} & \textbf{mF1}     & \textbf{wF1}     & \textbf{mF1}           \\ 
\hline
\multicolumn{13}{c}{\textbf{ViCGCN}}                                                                                                                                                                                                                                                                                                               \\ 
\hline
Performance       & \textbf{71.33}                    & \textbf{67.82}    & \textbf{72.08}                    & \textbf{68.12}   & \textbf{86.12}                    & \textbf{85.88}   & \textbf{93.11}                    & \textbf{76.12}    & \textbf{94.83}                    & \textbf{84.23}   & \textbf{91.02}   & \textbf{81.88}         \\ 
\hline
\multicolumn{13}{c}{\textbf{w/o GCN}}                                                                                                                                                                                                                                                                                                              \\ 
\hline
Performance       & 65.12                             & 63.23             & 71.13                             & 65.12            & 81.03                             & 79.53            & 90.12                             & 73.32             & 93.24                             & 82.96            & 88.72            & 79.12                  \\
Decrease          & $\downarrow$6.21                  & $\downarrow$4.59  & $\downarrow$0.95                  & $\downarrow$3.00 & $\downarrow$5.09                  & $\downarrow$6.35 & $\downarrow$2.99                  & $\downarrow$2.80  & $\downarrow$1.59                  & $\downarrow$1.27 & $\downarrow$2.30 & $\downarrow$2.76       \\ 
\hline
\multicolumn{13}{c}{\textbf{w/o PhoBERT}}                                                                                                                                                                                                                                                                                                          \\ 
\hline
Performance       & 52.32                             & 51.32             & 61.34                             & 58.42            & 79.63                             & 78.37            & 87.63                             & 64.32             & 88.32                             & 75.32            & 85.36            & 75.21                  \\
Decrease          & $\downarrow$19.01                 & $\downarrow$16.50 & $\downarrow$10.74                 & $\downarrow$9.70 & $\downarrow$6.49                  & $\downarrow$7.51 & $\downarrow$5.48                  & $\downarrow$11.80 & $\downarrow$6.51                  & $\downarrow$8.91 & $\downarrow$5.66 & $\downarrow$6.67       \\
\hline
\end{tabular}}
\end{table}

Our proposed method is considerably more effective than most current techniques for classifying text on social media. Ablation experiments were carried out on the proposed approach to prove the effectiveness of these two modules, PhoBERT and GCN. Table \ref{tab::Ablation} shows the ablation experiment results of the text classification module. For the model with GCN ablation, the experimental results are inferior to the model without ablation. While results of the \textit{w/o PhoBERT} model are not as good as those of the model with the contextualized pre-trained language model. The results of the ablation experiments demonstrate the effectiveness of the proposed importance of each module in general, as well as the combination of our proposed approach in particular. Our proposed approach, especially contextualized language models Integrated with graph neural networks, yield promising outcome for improving performance in further study. As a result, we conclude that all proposed modules are crucial in text classification on social media.

%% file: sections/conclusion.tex
\section{Conclusion and Future Work} 
\label{Conclusion}

In this study, we proposed a novel approach combining ViCGCN to take advantage of the powerful contextualized word representations learned by PhoBERT and leverage the graph structure of Graph Convolutional Networks (GCN). Moreover, we verified the impact of GCN on BERTology. The experimental results indicate that ViCGCN outperformed 13 powerful baseline models, including BERTology models, fused BERTology-GCN models, and state-of-the-art methods, across three benchmark social media datasets. Our proposed approach, ViCGCN, significantly improves up to 6.21\%, 4.61\%, and 2.63\% compared to the best Contextual Language Model and 2.38\%, 1.50\%, and 1.86\% compared to the best baseline model and previous studies on UIT-VSMEC, UIT-ViCTSD, and UIT-VSFC datasets, respectively. Additionally, our proposed ViCGCN model successfully addresses the challenge of imbalanced and noisy data in benchmark social media datasets. Finally, incorporating GCN as the final layer in BERTology models leads to a substantial improvement in performance, which claims the huge impact of GCN on Contextualized Language Models on text classification tasks.

Despite the outstanding results of our proposed approach, a practical text classification system that can be applied in real-world scenarios is necessary. Moreover, to improve the accuracy and robustness of our system, automatic pre-processing techniques for text normalization are required. This could include converting slang or informal language to the standard text, detecting and resolving spelling errors, or identifying and removing redundant information. Furthermore, we aim to experiment with other graph neural network models to compare their performance with the ViCGCN model, such as Graph Attention Networks (GATs) \cite{velickovic2018graph} and other graph-based models \cite{wu2020comprehensive}.